\documentclass{article}

\usepackage[preprint]{neurips_2026}
\usepackage[utf8]{inputenc} 
\usepackage[T1]{fontenc}    
\usepackage{hyperref}       
\usepackage{url}            
\usepackage{booktabs}       
\usepackage{amsfonts}       
\usepackage{nicefrac}       
\usepackage{microtype}      
\usepackage[dvipsnames]{xcolor}         
\usepackage{arydshln}

\usepackage{hyperref}
\hypersetup{
    colorlinks=true,    
    citecolor=citeblue, 
    linkcolor=red,      
    urlcolor=blue       
}

\definecolor{citeblue}{RGB}{0, 114, 189} 

\usepackage{graphicx}
\usepackage{booktabs}
\usepackage{xspace}
\usepackage{csquotes}
\usepackage{color, colortbl}
\usepackage{enumitem}
\usepackage{subcaption}
\usepackage[ruled,linesnumbered]{algorithm2e}

\usepackage{amsmath}
\usepackage{makecell}
\usepackage{multirow}
\usepackage{float}
\usepackage{amssymb}
\usepackage{utfsym}
\usepackage{diagbox}
\usepackage{dsfont}
\usepackage{cleveref}
\crefname{section}{Sec.}{Secs.}
\Crefname{section}{Section}{Sections}
\Crefname{table}{Table}{Tables}
\crefname{table}{Tab.}{Tabs.}
\crefname{algocf}{Alg.}{Algs.}
\Crefname{algocf}{Algorithm}{Algorithms}

\PassOptionsToPackage{numbers, compress}{natbib}

\newcommand{\para}[1]{\medskip\noindent\textbf{#1.}}

\definecolor{queryAction}{HTML}{DCEBFA}
\definecolor{queryStatus}{HTML}{F8D7DA}
\definecolor{queryWeather}{HTML}{DDF3E4}
\definecolor{queryAnimal}{HTML}{EEE2FA}

\newcommand{\eg}{e.g.,\xspace}

\title{Test-Time Coverage: Test-Conditioned Data Curation for Deployment-Aware Learning}

\author{%
    Nadine Chang* \\
    NVIDIA \\
    \texttt{nadinec@nvidia.com} \\
    \And
    Maying Shen* \\
    NVIDIA \\
    \texttt{mshen@nvidia.com} \\
    \And
    Shizhe Diao \\
    NVIDIA \\
    \texttt{sdiao@nvidia.com} \\
    \And
    Jialiang Wang\\
    NVIDIA \\
    \texttt{jialiangw@nvidia.com} \\
    \And
    Jingde Chen \\
    NVIDIA \\
    \texttt{joshchen@nvidia.com} \\
    \And
    Thomas Breuel \\
    NVIDIA \\
    \texttt{tbreueul@nvidia.com} \\
    \And
    Pavlo Molchanov \\
    NVIDIA \\
    \texttt{pmolchanov@nvidia.com} \\
    \And
    Rafid Mahmood \\
    NVIDIA \& University of Ottawa \\
    \texttt{rmahmood@nvidia.com} \\
    \And
    Jose M. Alvarez \\
    NVIDIA \\
    \texttt{josea@nvidia.com} \\    
}

\begin{document}

\maketitle

\begin{abstract}

Deployed AI systems are often trained from broad candidate data pools, necessitating data curation towards the deployment test distribution. However, standard data curation methods score training-side criteria rather than directly optimizing deployment match. We introduce \textbf{TTCov (Test-Time Coverage)}, a data-level test-conditioned curation method that uses test-side information before training instead of updating model weights at inference. TTCov decomposes deployment-conditioned curation into coverage and distribution. To represent coverage, it builds a task \emph{Atlas}, a collection of LLM-based atomic propositions (APs) describing deployment-relevant concepts, seeded from open task knowledge and expanded with unmatched APs extracted from unlabeled deployment samples. To represent distribution, it instantiates the matched deployment APs with their frequencies, yielding a \emph{Knowledge Atlas} (K-Atlas) that operationalizes the deployment distribution as a curation target. TTCov then selects a budgeted training set whose deployment APs distribution approximates this target. We apply TTCov towards autonomous driving (AD), keeping adaptation off the inference path while selecting data with greater deployment-relevant coverage, closer K-Atlas matching, and stronger downstream end-to-end driving performance than data-curation baselines, including seamless adaptability to novel domains via city-to-city expansion. Code available at \url{https://github.com/NVlabs/TTCov}.

\end{abstract}

\section{Introduction}

Deployed AI systems often have access to broad candidate data pools for training the underlying model. 
For example, physical AI systems such as autonomous driving (AD) and robotics can collect large amounts of sensor data from production fleets, and large-scale benchmarks covering the breadth of user scenarios~\citep{Grzywaczewski2017NvidiaScale, Coren2025TeslaData, caesar2020nuscenes, sun2020waymo, liao2022kitti360}. 
Because training on every collected example is computationally infeasible, data curation policies seek to mine the data pool for a training subset that optimizes downstream test performance \citep{pattnayak2024survey, liu2024survey}.

Data curation methods often score samples and subsets from the data pool on training-side criteria such as diversity, representativeness, uncertainty, redundancy, or utility~\citep{sener2018coreset, coleman2020selection, shen2025sse}. 
These criteria can produce compact and useful training subsets, but they do not directly ask whether the selected dataset matches the distribution the model will encounter after training. 
To mitigate the potential mismatch, unlabeled deployment-side information can be used to match the curated training distribution to the test distribution that the deployed system will encounter~\citep{chang2026position}.

Statistical learning theory assumes a fixed test distribution and minimizes training error as a proxy for test error~\citep{vapnik2013nature}. In contrast, training and testing sets for deployed AI systems are often constructed and evolved by hand with human-curated datasets that emphasize specific critical evaluation conditions \citep{cao2025pseudo}. 
However, human curation does not scale to the shifting deployment conditions for AI technology products. 
For these situations, transductive learning methods that condition on unlabeled deployment-side data have shown improved performance when test conditions differ from the original training distribution~\citep{joachims1999transductive, bennett1999semi, zhu2003semi, zhou2004learning, belkin2006manifold}. More recent test-time training and test-time adaptation techniques instantiate this premise by using test data online to update model weights at inference~\citep{sun2020ttt, wang2021tent, liu2021ttt++, niu2022efficient,hardt2024test}.
Although these methods have been shown to be effective in physical AI settings \citep{sima2025centaur}, online model adaptation in these applications multiplies inference latency and complicates model validation under real-time use conditions such as safety-critical human interaction~\citep{wang2022cotta, niu2022efficient}. These methods also tune the model per test sample, which scales poorly with deployment volume.

\begin{figure}
    \centering
    \includegraphics[width=1.0\linewidth]{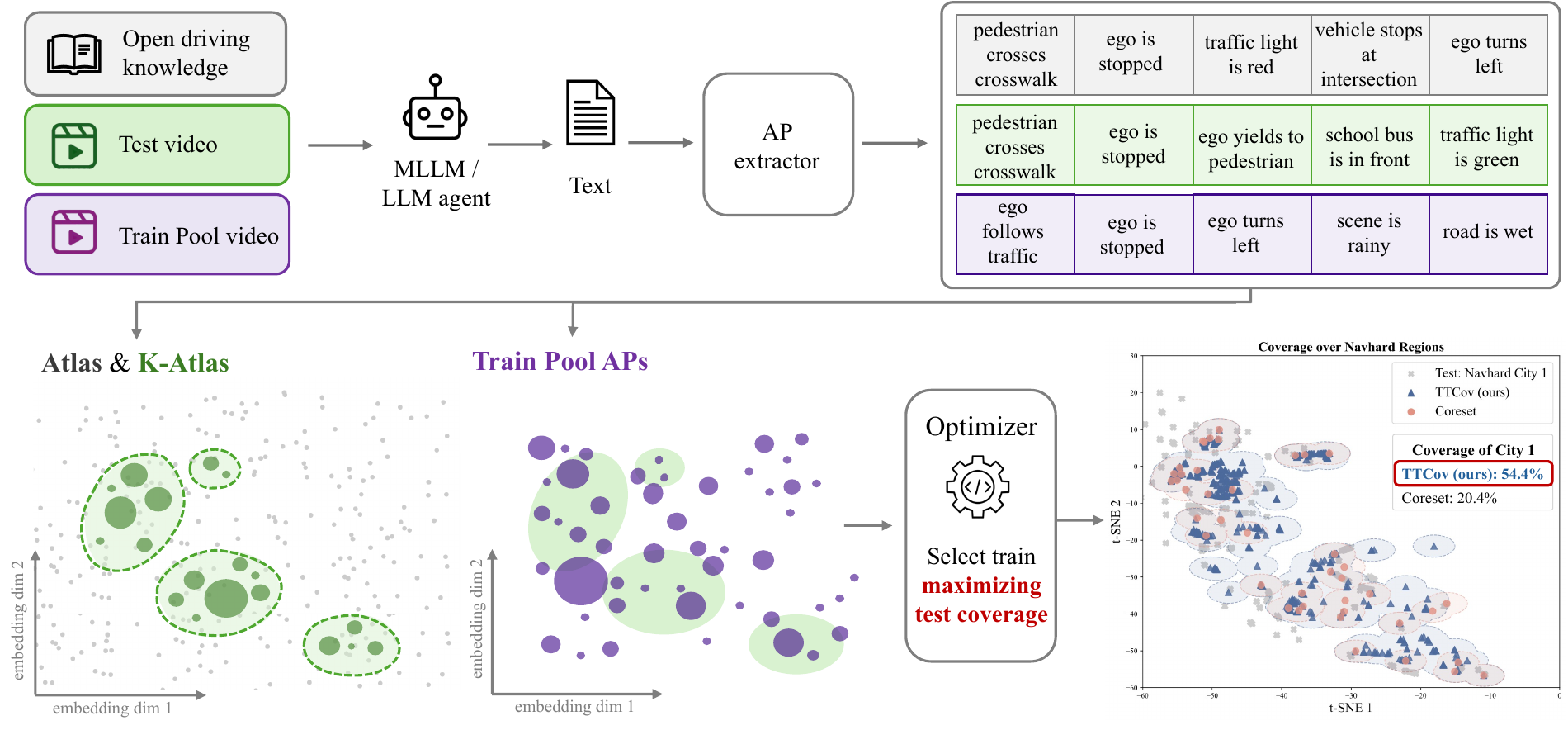}
    \caption{
    The Atlas consists of a collection of LLM-based atomic propositions (APs) describing deployment-relevant concepts. [Top] It is seeded from open task knowledge (grey) and expanded with any new APs extracted from test samples (green). [Bottom] To represent distribution, the frequency of test APs is calculated, yielding K-Atlas. TTCov aims to curate a training dataset that contains relevant test APs, matching the target test distribution and, consequently, maximizing test coverage, as shown in lower right plot.  
    }
    \label{fig:teaser}
    \vspace{-4mm}
\end{figure}

In this work, we introduce \textbf{TTCov (Test-Time Coverage)}, a data-level test-conditioned curation method that adapts the training distribution before training rather than adapting model weights at inference (~\cref{fig:teaser}). TTCov breaks deployment-conditioned curation into two problems: (i) coverage, which asks what deployment-relevant content the training data should include, and (ii) distribution, which asks how frequently that content should appear. To represent coverage, TTCov builds an interpretable \emph{task Atlas}: a collection of \emph{LLM-based \textbf{\underline{a}}tomic \textbf{\underline{p}}ropositions} (APs) that describe deployment-relevant concepts such as scenarios, agents, conditions, and behaviors. The Atlas is seeded from open task knowledge and expanded with APs extracted from unlabeled deployment-side samples when those samples contain concepts not already represented. To represent distribution, TTCov instantiates the matched deployment-side APs with their frequencies, yielding the \emph{Knowledge Atlas} (K-Atlas), an operational curation target for the deployment distribution. Our data curation pipeline then seeks to select a training dataset whose own K-Atlas, deployment relevant AP distribution, approximates the deployment K-Atlas.

We implement TTCov on autonomous driving, a safety-critical physical AI domain where the deployment distribution is available at training time via driving logs, target operational design domains, sensor stacks, and route specifications. Here, the Atlas is seeded from open driving knowledge such as taxonomies extracted from large language models (LLMs) \citep{comanici2025gemini,achiam2023gpt,yang2025qwen3}, and it expands when test-derived APs do not match existing entries~\citep{caesar2020nuscenes}. We demonstrate the effectiveness of TTCov in variable deployment conditions through city-to-city expansion, where the training set must be re-curated for a new city. Against the breadth of data curation techniques, TTCov selects data that covers more deployment-relevant APs, better matches the target K-Atlas distribution, and improves end-to-end (E2E) autonomous driving performance. Our contributions are:

\begin{enumerate}
    \item We propose TTCov, a data-level method for test-conditioned curation that adapts the training dataset to the unlabeled deployment distribution before model training.
    \item We introduce the Atlas and K-Atlas as an interpretable coverage representation and frequency-weighted target distribution for deployment-conditioned data curation.
    \item We formulate TTCov as budgeted K-Atlas matching and solve it with a greedy selection procedure over candidate training samples.
    \item We provide a monotonic Atlas evolution mechanism for changing deployment conditions, instantiated through city-to-city domain expansion in autonomous driving.
    \item We evaluate TTCov with coverage, distribution-matching, and downstream E2E autonomous driving metrics against data curation baselines.
\end{enumerate}

\section{Related Works}
\vspace{-2mm}
\textbf{Data curation} selects subsets of a candidate pool by training-side criteria.
Coverage and diversity based methods aim to span the pool's feature or gradient space~\citep{sener2018coreset, ash2019deep, coleman2020selection, smith2023prediction}.
Deduplication methods remove visually or semantically duplicate samples, often via CLIP-based similarity~\citep{abbas2023semdedup, clip2021, kang2025adadedup, slyman2024fairdedup, schuhmann2022laion}. Uncertainty active learning selects samples on which the model is least confident~\citep{lewis1994heterogeneous, joshi2009multi, ren2021survey}. Utility and difficulty based methods score samples by their effect on training or model behavior~\citep{paul2021deep, toneva2018an, shen2025sse, dimlioglu2026scaling}, and scaling-aware analyses show that pruning can improve scaling-law behavior~\citep{sorscher2022beyond, goyal2024scaling}. All these criteria are defined entirely based on training data. They do not consider whether the selected subset matches the distribution the deployed model will encounter~\citep{chang2026position}, which can cause selected data to be distributional misaligned with frequently shifting deployment distributions in safety-critical physical AI. Unlike these methods, TTCov performs test-conditioned curation method to ensure deployment distributional alignment.

\textbf{Transductive Learning} conditions on test-side information to improve predictions on specific test instances. Introduced in~\cite{vapnik2013nature}, specific transductive problem is argued to be easier to solve than the general inductive one. Classical methods include transductive SVMs using unlabeled test points to find low-density separators~\citep{joachims1999transductive, bennett1999semi}, graph-based label propagation spreading labels through a similarity graph that has the test set~\citep{zhu2003semi, zhou2004learning}, and manifold regularization~\citep{belkin2006manifold}. These methods assume a fixed labeled test set, work only on prediction stage, and do not scale to modern models or evolving deployment distributions. Modern works bring the transductive principle to model adaption: test-time training (TTT) adapts model weights at inference via a self-supervised auxiliary task~\citep{sun2020ttt, liu2021ttt++,hardt2024test}. Test-time adaptation (TTA) replaces the auxiliary task with a minimized objective in the test stream~\citep{wang2021tent, niu2022efficient, wang2022cotta, niu2023sar, zhang2022memo}. While modern model adaptation has shown effectiveness for E2E AD~\citep{sima2025centaur}, their online updates present unscalable per-sample compute and latency.

\textbf{End-to-end Autonomous Driving} trains planning models that directly use raw sensor data to directly predict trajectories. Modern approaches use imitation learning to mimic expert ground-truth behavior~\cite{codevilla2018end, chitta2023transfuser, jiang2023vad, chen2024vadv2, li2024hydra, li2025ztrs} and evaluate with both open and closed-loop metrics. However, open-loop metrics have been proven to be misaligned with true closed-loop driving quality~\cite{li2024ego, dauner2023parting}. In an effort to translate to real-world driving, approaches now evaluate on closed-loops simulation benchmarks~\cite{dauner2024navsim, cao2025pseudo}, which we similarly evaluate in our work.

\vspace{-2mm}
\section{TTCov}
\label{sec:method}

In this section, we formulate TTCov, a framework for structuring knowledge relevant for deployment-conditioned data curation before model training. TTCov separates the curation problem into two stages: coverage, which asks what deployment-relevant knowledge should the training data include, and distribution matching, which asks how frequently should this knowledge appear. Below, we describe TTCov as instantiated on an AD task.

\begin{figure}[t]
    \centering
    \includegraphics[width=1.0\linewidth]{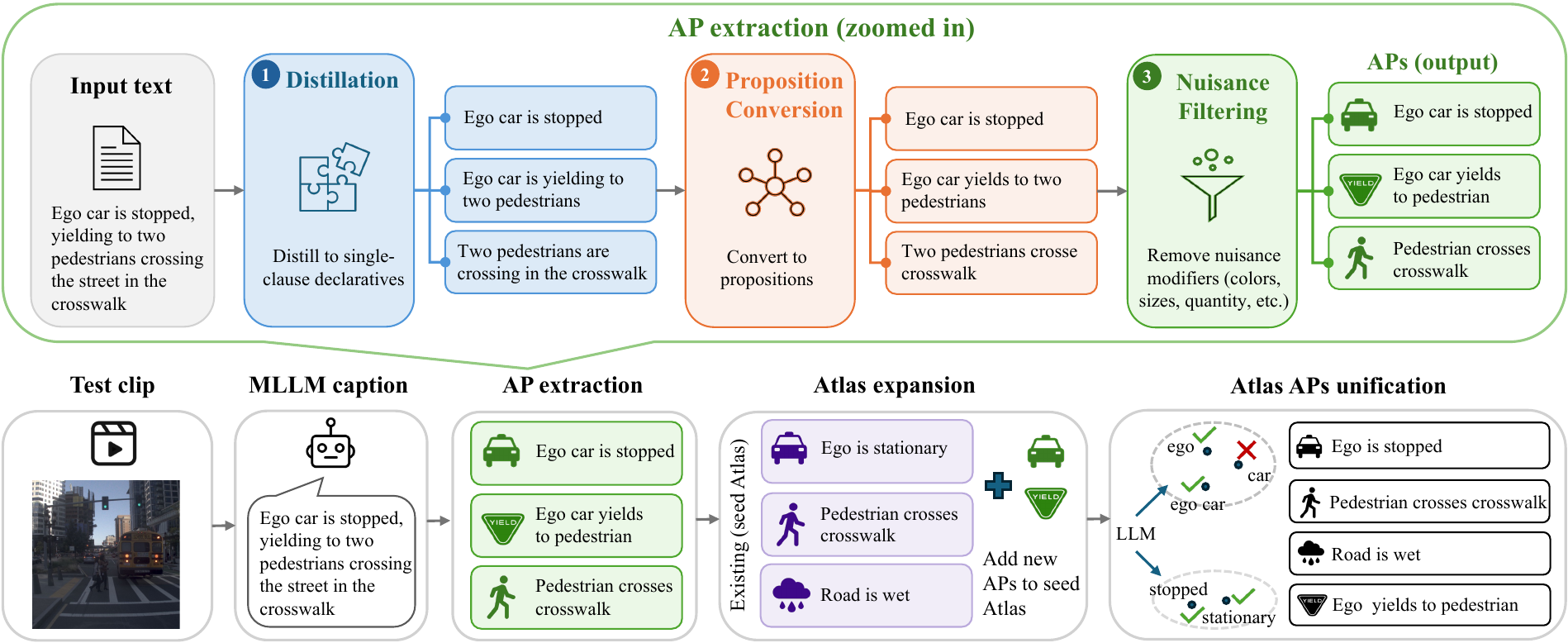}
    \caption{TTCov Overview. Top: The AP extraction process transforms complex descriptions into clean, atomic propositions and forms the list of APs. Bottom: Workflow for generating the K-Atlas. Video test clips are captioned and converted into APs, which are used to expand the seed Atlas. The final unification stage uses an LLM to merge similar phrases (e.g., ``stationary'' and ``stopped''), ensuring a consistent phrasing.}
    \label{fig:ttcov_overview}
    \vspace{-2mm}
\end{figure} 

\subsection{Covering deployment knowledge via an \emph{Atlas}}

We define an Atlas as a set of LLM-based atomic propositions (APs) that cover all knowledge required to complete the task. Each AP is a fundamental declarative sentence unit that describes concepts such as scenarios, agents, conditions, and behaviors (e.g., \enquote{A pedestrian crosses crosswalk}, \enquote{A vehicle yields at an intersection}, \enquote{The road is wet}. Figure \ref{fig:ttcov_overview} summarizes the Atlas creation process.

For our AD setting, APs are extracted from input plain text obtained from two knowledge sources. To seed an initial open-world knowledge, we note that LLMs trained on internet-scale data can generate strong structured ontologies for general tasks. We seed an initial Atlas by prompting an LLM to construct an ontology of driving conditions. 
We then augment this knowledge base with deployment test data. Here, an MLLM extractor translates each recorded driving video in the deployed test set into dense captions, answering a series of questions about the corresponding scenarios, agents, and behaviors. Our complete input knowledge base is the set of input texts from the initial seed Atlas and captions from the deployment data. We provide full prompt details for both extraction steps in~\cref{app:atlas}.

All input plain text is converted into APs via a three stage process (see top of~\cref{fig:ttcov_overview} for an example input text being parsed into APs). First, we extract a compact semantic description that preserves the salient concepts present in the input text into single‑clause declarative sentences. 
Then, we further transform distilled descriptions into a set of atomic noun-anchored propositions (i.e., subject-verb-object or subject-attribute). 
Finally, each proposition may still contain non-essential modifiers or lexical noise. As a result, we apply de-noising filtering by removing all nuisance modifiers. This yields a list of APs relevant to our task.

APs obtained from multiple sources (e.g., seed Atlas, different driving videos) may have redundancy in equivalent concepts with slightly different phrasing. Consequently, we unify the list of APs into our Atlas of unique concepts. We perform this unification in a two-stage approach. First, we cluster all the nouns and verbs using sentence embeddings. Each cluster then contains semantically similar phrases. Then for each cluster, we prompt an LLM to verify if all cluster phrases share the same meaning, identify any outlier phrases, and pick one phrase to replace all equivalent entries. This de-duplication process reveals the Atlas.

Finally, we define an AP distribution as the Knowledge-Atlas (K-Atlas). To instantiate the K-Atlas in AD, we review the set of original APs from each video of the deployment test set. Within a given data sample, we perform pairwise matching for each AP against the set of APs in the Atlas using Faiss \citep{douze2025faiss}. Then, let $p^\star$ be the K-Atlas distribution vector, whose elements represent an importance frequency of each AP $L$ via:

\begin{equation}\label{eq:p_star_L}
    p^\star(L) := \frac{\#\{ \text{test samples whose APs include } L \}}{\sum_{L' \in \mathrm{Atlas}} \#\{ \text{test samples whose APs include } L' \}}.
\end{equation}

\begin{algorithm}[t]
\DontPrintSemicolon
\KwIn{Candidate pool $\mathcal{C}$, target K-Atlas $p^\star$, budget $B$, per-sample AP sets $\{\mathcal{L}(x)\}_{x \in \mathcal{C}}$}
\KwOut{Curated subset $S^\star$ with $|S^\star| = B$}
$S \leftarrow \emptyset$\;
\While{$|S| < B$}{
    \ForEach{$x \in \mathcal{C} \setminus S$}{
        $\hat{p}_{S \cup \{x\}} \leftarrow$ K-Atlas of $S \cup \{x\}$ over atlas APs\;
        $g(x) \leftarrow D_{\mathrm{KL}}\!\left(p^\star \,\|\, \hat{p}_{S \cup \{x\}}\right)$\;
    }
    $x^\star \leftarrow \arg\min_{x \in \mathcal{C} \setminus S} g(x)$\;
    $S \leftarrow S \cup \{x^\star\}$\;
}
\Return{$S^\star \leftarrow S$}\;
\caption{Greedy KL coverage for TTCov curation.}
\label{alg:greedy}
\end{algorithm}

\subsection{Curating a training set by targeting the \emph{K-Atlas} test distribution}

Given the set of task-relevant knowledge, our goal is to curate a training dataset conditioned on a deployment AP distribution. Let $\mathcal{C}$ be a candidate data pool of samples, e.g. driving videos. For each sample $x \in \mathcal{C}$, we apply the same MLLM-based extraction procedure to obtain a corresponding set of APs. To ensure that these APs are unified in the same vocabulary as our K-Atlas, we perform pairwise matching of these APs against the APs in our Atlas.

We seek to construct a curated subset $\mathcal{S} \subset \mathcal{C}$ with a budget $| \mathcal{S} | = B$ samples. In TTCov, we optimize for a subset whose distribution matches the target K-Atlas. Specifically for a given subset $\mathcal{S}$, let $\hat{p}_\mathcal{S}$ be the distribution vector whose elements represent the importance frequency of each AP in the set $\mathcal{S}$, analogous to \eqref{eq:p_star_L}. Then, the goal fo TTCov is to optimize
\begin{equation} \label{eq:ttcov_opt_problem}
    \mathcal{S}^\star := \arg\min_{\mathcal{S} \subset \mathcal{C} ,\; |\mathcal{S}| = B} D_{\mathrm{KL}}\!\left(p^\star \,\|\, \hat{p}_\mathcal{S}\right).
\end{equation}

The above problem \eqref{eq:ttcov_opt_problem} is a combinatorial optimization problem over a convex objective. Considering large candidate data pools increase the scale of the problem, we apply a greedy iterative algorithm by iteratively picking samples whose APs maximally reduce the current KL gap. Algorithm \ref{alg:greedy} summarizes our approach. 

\section{TTCov Analysis}
\label{sec:coverage}
\vspace{-1mm}

\para{LLMs and MLLMs} Unless stated otherwise, we use Gemini 2.5 Pro~\cite{comanici2025gemini} and Qwen 3 Embedding~\cite{yang2025qwen3} for all relevant calls in Atlas and K-Atlas creation. 
\vspace{-1mm}

\subsection{Atlas Ablations: Why LLM atomic propositions?}
\vspace{-1mm}
A natural starting point for our Atlas is a knowledge graph (KG), and recent work has shown that SOTA methods can generate KGs directly from large text corpus~\citep{mo2025kggen}. Starting from our seed Atlas, we use KG Gen to extract a knowledge graph. However, whole complex sentences collapse into a single triplet, losing shared atomic structure across examples. Consider two sentences: 1) \textit{``The car is slowly nudging to the left and stopping because there are three people crossing Broadway street and a car is parked on the street''} contains four key parts (nudging left, stopping, three people crossing, car parked), while 2) \textit{``The car is stopping because a person is crossing the avenue''} contains two (stopping, person crossing). These captions share key concepts --- stopping, people crossing --- but KG Gen treats them as entirely separate. We include a list of examples in~\cref{app:atlas}.

To capture shared similarity at the atomic level, we move away from KGs to atomic breakdowns of captions via LLMs. Each caption decomposes into basic atomic propositions (APs) across scenarios, actions, etc., \eg \textit{``Car turn left'', ``Person is crossing street'', ``Three people crossing street''}. More examples in~\cref{app:atlas}. However, modifiers like street names (\textit{Broadway}), adjectives (\textit{three}), synonyms (\textit{avenue} vs. \textit{street}), and plurals (\textit{cars} vs. \textit{car}) keep these APs separate. We note that none of these modifiers are critical for driving. We do not want to hit any number of people crossing any street, regardless of what color shirt they may be wearing. Thus, we add a final de-noising and phrasing unification stage to remove distracting modifiers and unify near-identical APs. The final result is our Atlas: a compact, deduplicated representation of driving knowledge in atomic form.

\vspace{-1mm}
\subsection{Atlas Coverage}
\label{sec:coverage_city}
\para{Datasets}
We use OpenScene trainval~\cite{openscene2023} split as our training data pool, which provides a rich diversity of driving recordings. The dataset contains driving session clips lasting from 30 seconds to 50 minutes. To better capture the actions during driving and align with industry standards~\cite{dimlioglu2026scaling}, we segment each session into fixed-length 10-second virtual clips (20 frames at 2Hz) with a sliding window. Adhering to the Navsim~\cite{dauner2024navsim} framework configuration, each clip contains a 3-frame historical buffer and a 10-frame future horizon. The remaining frames are added to the training set upon virtual clip selection. Further details are available in~\cref{app:data}. We use Navhard~\cite{cao2025pseudo} as the test dataset. Unlike standard test sets, Navhard isolates high-difficulty scenes, providing a more stringent metric for evaluating the planning stability and safety of the proposed model in edge-case environments. For our city expansion experiments, we iteratively add the cities in Navhard in the following order: Pittsburgh, Las Vegas, Singapore, and Boston. 

\para{Baselines} We compare TTCov against two baseline data selection strategies. Both methods aim to ensure that a model trained on the subset performs as closely as possible to a model trained on the entire pool. Coreset~\cite{sener2018coreset} selects data points from the pool that maximize diversity and coverage over the entire training feature space. SSE~\cite{shen2025sse} clusters data points from the pool and maximize the diversity by removing semantically repetitive samples. 

\begin{table}[t]
  \centering
  \caption{TTCov's coverage metrics for multi-round city evolution. For each round, we add a new test city and report the number of
  selected training points, whose embedding is within 0.15 cosine dist of any test point
  (Num. Selected Train, NN @ 0.15), and Maximum Mean Discrepancy (MMD).}
  \label{tab:city_metrics}
  \begingroup
  \small
  \setlength{\tabcolsep}{3pt}
  \renewcommand{\arraystretch}{1.15}
  \begin{tabular*}{\textwidth}{@{\extracolsep{\fill}}c@{\quad\quad}cccc@{\quad\quad}cccc@{}}
    \toprule
    & \multicolumn{4}{c}{Num. Selected Train, NN @ 0.15 ($\uparrow$)}
    & \multicolumn{4}{c}{MMD($\downarrow$)} \\
    \cmidrule(r{0.75cm}){2-5} \cmidrule(l){6-9}
    Method & Round 1 & Round 2 & Round 3 & Round 4
           & Round 1 & Round 2 & Round 3 & Round 4 \\
    \midrule
    Coreset & 42 & 36 & 19 & 62 & 0.13 & 0.10 & 0.20 & 0.13 \\
    SSE     & 46 & 49 & 17 & 68 & 0.13 & 0.11 & 0.20 & 0.13 \\
    \textbf{TTCov (ours)}    & \textbf{732} & \textbf{355} & \textbf{47} & \textbf{228}
            & \textbf{0.08} & \textbf{0.06} & \textbf{0.14} & \textbf{0.10} \\
    \bottomrule
  \end{tabular*}
  \endgroup
\end{table}
\begin{figure}[t]
    \centering
    \includegraphics[width=1.0\linewidth]{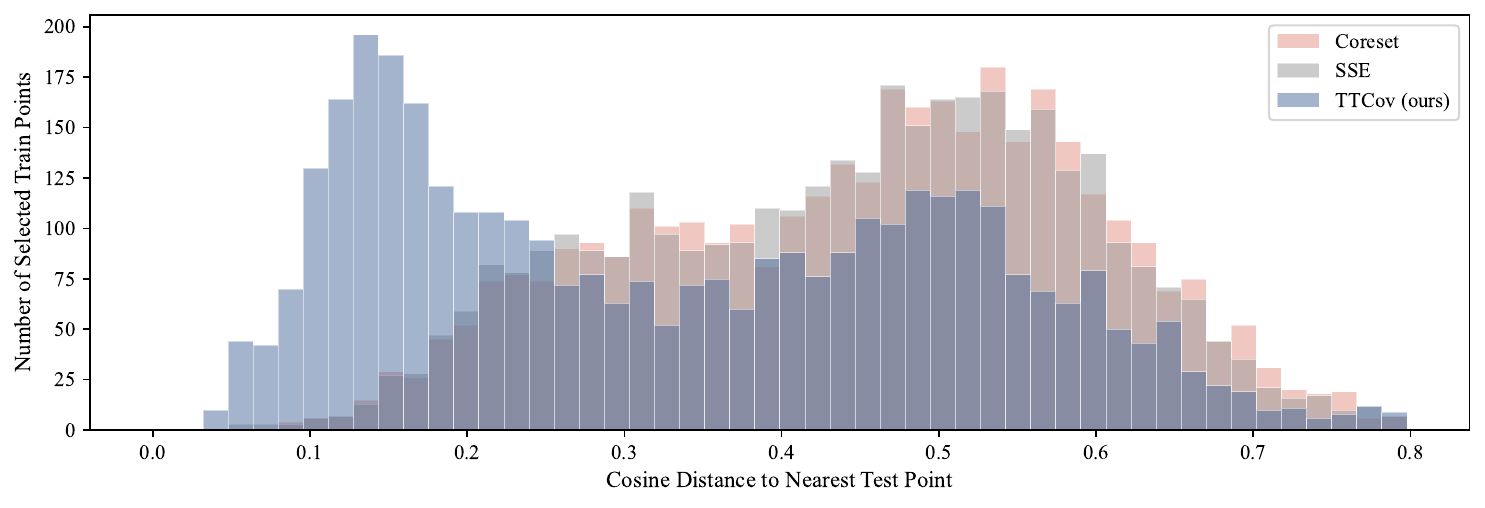}
    \caption{TTCov's selected data coverage over first test city. We show the histogram of the number of selected data points, whose embeddings are within a certain distance from any nearest test points. Given only one target test city, TTCov selects more data points closer to test points (see left of fig).
    }
    \label{fig:city_cover}
    \vspace{-2mm}
\end{figure}
\para{Atlas can adapt to dynamic test sets and maximize coverage}
As operational task goals shift, test-conditioned curation must address dynamically evolving test sets. In AD, this manifests as a continuously growing set of corner cases existing datasets often fail to cover and is critical to identify. While some existing methods~\citep{smith2023prediction} attempt to address curation by matching train labels, they require discrete labels for tasks like classification. However, E2E planning in AD consists of continuous trajectories over a given time span. TTCov does not require such discrete labels. We evaluate TTCov ability to address this gap with a city-to-city domain adaptation setting, a common goal for AD. TTCov addresses this evolving requirements by matching new test city data against its existing Atlas and updating the K-Atlas for the new city. Crucially, TTCov retains previously selected data from prior cities while adding only the most relevant samples for the new domain, ensuring efficient expansion without redundancy. 

To evaluate data coverage efficacy, we use established coverage metrics, the Nearest Neighbor (NN) distance between closest two points in two datasets~\cite{hardt2024test} and Maximum Mean Discrepancy (MMD) with a radial basis function. As shown in~\cref{tab:city_metrics}, TTCov’s adaptation strategy maximizes test set coverage, yielding significantly more selected training data points close to existing test data points (NNs) and lower MMD than competing baselines. Notably, this advantage grows with domain evolution, where TTCov continues to gain coverage with each additional city.~\cref{fig:city_cover} visualizes the selected subset for the initial city, counting the amount of selecting data point within a threshold proximity of a nearest test point. Compared to other methods, TTCov selects a far denser concentration of points relevant to the test distribution. These metrics collectively highlight TTCov’s flexibility, robustness, and ability for continuous adaptation across diverse domains. Additional results in ~\cref{app:atlas}.
\section{Experiments}
\label{sec:exps}
We evaluate TTCov on Navhard, containing challenging scenarios from Nuplan, across various budgets and report E2E performance, EPDMS. Below details budgets, models, and metrics.

\vspace{-1mm}
\para{Budgets}
We use the size of Navtrain, a manually curated dataset from OpenScene trainval split, as our baseline budget ($1\times$) for all studies. To evaluate the scalability of our approach, we further conduct a budget sweep across varying magnitudes, specifically at $0.5\times$, $0.75\times$, $1.25\times$ and $1.5\times$ the baseline budget $B$ for our main results.

\vspace{-1mm}
\para{Models and Metrics} We evaluate data selection methods by training Latent Transfuser (LTF)~\cite{chitta2023transfuser}, a baseline planner integrated into the NAVSIM framework~\cite{dauner2024navsim,cao2025pseudo}. All models are trained for 100 epochs on eight NVIDIA A100 GPUs using the default NAVSIM training configuration. Trained checkpoints are evaluated on the NAVSIM-v2 benchmark~\cite{cao2025pseudo}, and we report EPDMS scores on the \texttt{navhard\_two\_stage} split. NAVSIM-v2 extends NAVSIM~\cite{dauner2024navsim} with a two-stage pseudo-simulation protocol that augments real driving recordings with synthetic novel views to better approximate closed-loop evaluation. The \texttt{navhard} split is its most challenging curated evaluation subset, containing diverse and operationally difficult traffic and road geometries. For each data selection training split, we train models with three random seeds and the average EPDMS is reported, std. dev. in~\cref{app:metrics}.

\begin{table}[h]
  \centering
  \caption{TTCov outperforms other curation baselines on Navhard. We report EPDMS and its ratio to the original manually curated dataset Navtrain (oracle), where 1 indicates parity to Navtrain.}
  \label{tab:main_tab}
  \begingroup
  \footnotesize
  \setlength{\tabcolsep}{1pt}
  \renewcommand{\arraystretch}{1.15}
  \begin{tabular*}{\textwidth}{@{\hspace{1.2em}\extracolsep{\fill}}cccccc@{\hspace{1.2em}}}
    \toprule
    & \multicolumn{5}{c}{Budget} \\
    \cmidrule(lr){2-6}
     & 0.5x & 0.75x & 1x & 1.25x & 1.5x \\
    \cmidrule(lr){2-6}
    Method & \scriptsize EPDMS($\uparrow$) | Rel.($\uparrow$) & \scriptsize EPDMS | Rel. & \scriptsize EPDMS | Rel. & \scriptsize EPDMS | Rel. & \scriptsize EPDMS | Rel. \\
    \midrule
    Navtrain (oracle)         & -- & -- & 24.70 | 1.00 & -- & -- \\
    Random           & 18.95 | 0.77 & 18.76 | 0.76 & 20.15 | 0.82
                     & 21.95 | 0.89 & 22.95 | 0.93 \\
    Coreset          & \textbf{20.77 | 0.84} & 21.48 | 0.87
                     & 23.63 | 0.96 & 24.55 | 0.99 & 25.89 | 1.05 \\
    SSE              & 18.44 | 0.75 & 22.29 | 0.90 & 23.45 | 0.95
                     & 24.77 | 1.00 & 26.03 | 1.05 \\
    \textbf{TTCov (ours)}     & 20.62 | 0.83 & \textbf{23.00 | 0.93}
                     & \textbf{24.42 | 0.99} & \textbf{25.49 | 1.03}
                     & \textbf{26.40 | 1.07} \\
    \bottomrule
  \end{tabular*}
  \endgroup
\end{table}

\subsection{Main Results}
\vspace{-2mm}

\para{Navhard} To evaluate the efficacy of TTCov, we select subsets from an unlabeled training pool (OpenScene) and evaluate them on the E2E task in Navhard, a collection of the most challenging scenes from NAVSIM. Notably, the original Navtrain dataset was manually curated using ground-truth labels to filter out annotation errors and non-trivial solutions~\cite{dauner2024navsim}. As such, we treat Navtrain as an oracle upper bound. In \cref{tab:main_tab}, we report both absolute EPDMS scores and the relative gains compared to the oracle across data budgets. For brevity, PDM submetrics are in~\cref{app:metrics}. TTCov demonstrates consistent performance gains over all baselines, including the SOTA method, SSE \cite{shen2025sse}. As the budget approaches $B=1$, TTCov approaches oracle performance more than other methods. Furthermore, at budgets exceeding the original Navtrain size, TTCov outperforms all baselines and the oracle itself. Crucially, at $B=1$, TTCov automatically curates a dataset that achieves performance parity with Navtrain without requiring expensive human annotations or manual filtering.

\para{TTCov for city evolution} 
Beyond the city evolution ablation study in \cref{sec:coverage_city}, we train the E2E model on our selected data for each city added. Notably, TTCov proves truly adaptive; it only requires data selection for each new environment without re-selecting data for previous cities. As shown in \cref{tab:city_tab}, TTCov significantly outperforms all baselines for City 1 in the initial round. In Round 2, TTCov not only outperforms others in the newly introduced City 2 but also preserves its performance lead in City 1. This highlights TTCov’s ability to continuously adapt to new domains without suffering from catastrophic forgetting. Ultimately, TTCov far outpaces competing methods across both cities, demonstrating its general robustness and stability in evolving environments.

\begin{table}[t]
  \centering
  \caption{City to city performance. TTCov naturally extends Atlas to
  new cities without re-curation and improves EPDMS($\uparrow$) in both previous and current targeted cities.}
  \label{tab:city_tab}
  \begingroup
  \small
  \setlength{\tabcolsep}{16pt}
  \renewcommand{\arraystretch}{1.15}
  \vspace{1.0ex}

  \begin{tabular}{c@{\quad\quad}c@{}p{3.5em}@{}>{\centering\arraybackslash}p{3.5em}>{\centering\arraybackslash}p{3.5em}>{\centering\arraybackslash}p{3.5em}}
    \toprule
    & Round 1 & & \multicolumn{3}{c}{Round 2} \\
    \cmidrule(l{0em}r{0em}){2-2} \cmidrule(l{0em}r{0.2em}){4-6}
    Method & City 1 & & City 1 & City 2 & City 1+2 \\
    \midrule
    Coreset       & 15.72 & & 20.29 & 29.69 & 22.80 \\
    SSE           & 15.78 & & 19.91 & 31.18 & 22.18 \\
    \textbf{TTCov (ours)}  & \textbf{20.23} & & \textbf{21.94} & \textbf{38.11}
                  & \textbf{26.39} \\
    \bottomrule
    
  \end{tabular}
  \endgroup
  \vspace{-3mm}
\end{table}

\begin{figure}[t!]
  \centering
  
  \begin{minipage}[t]{0.4\textwidth}
    \centering
    \vspace{0pt} 
    \includegraphics[width=\textwidth]{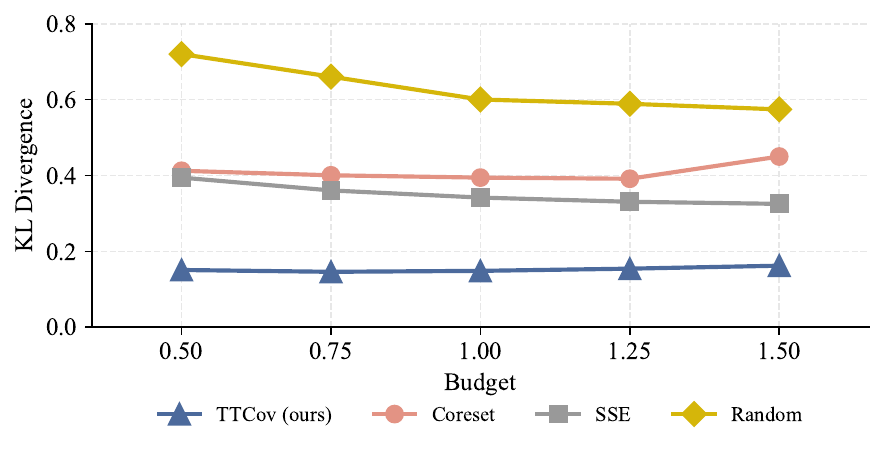}
    \caption{KL div to K-Atlas distribution.}
    \label{fig:kl_ours}
  \end{minipage}
  \hfill
  \begin{minipage}[t]{0.55\textwidth}
    \centering
    \vspace{4mm} 
    \captionof{table}{Distribution metrics at budget 1x.} 
    \label{tab:dist_metrics}
    \scriptsize 
    \setlength{\tabcolsep}{3pt}
    \begin{tabular}{lccccc}
   
      \toprule
      Method & KL Div.($\downarrow$) & JS Dist.($\downarrow$) & Hell.($\downarrow$) & Cosine($\uparrow$) & Cov.($\uparrow$) \\
      \midrule
      Random & 0.60 & 0.31 & 0.32 & 0.78 & 0.67 \\
      Coreset & 0.34 & 0.23 & 0.24 & 0.90 & 0.72 \\
      SSE    & 0.34 & 0.23 & 0.24 & 0.90 & 0.72 \\
      \textbf{TTCov (ours)}  & \textbf{0.15} & \textbf{0.14} & \textbf{0.14} & \textbf{0.98} & \textbf{0.80} \\
      \bottomrule
    \end{tabular}
  \end{minipage}
  \vspace{-2mm}
\end{figure}

\para{TTCov selects data with closest distributional match to test} 
By targeting our K-Atlas test distribution comprised of relevant APs, TTCov curates a training set through an optimizer that minimizes the distributional distance between the selected data and the target K-Atlas. This alignment is visualized in \cref{fig:kl_ours}, which shows that TTCov's selected data distribution is significantly closer to K-Atlas than other methods. Crucially, our results confirm a correlation between distributional proximity and model performance. The closer the selected dataset aligns with the K-Atlas test distribution, the higher the resulting performance. For example, as shown in~\cref{tab:main_tab}, TTCov achieves the highest EPDMS score alongside the lowest KL divergence, whereas Coreset and SSE exhibit both higher KL divergence and correspondingly poorer EPDMS performance. Lastly, this trend remains consistent across all several distribution metrics. As shown in \cref{tab:dist_metrics}, TTCov consistently achieves the closest distributional proximity regardless of the metric evaluated.

\vspace{-1mm}
\subsection{Ablations}
\vspace{-1mm}

\para{Optimizer Ablations} We study the various effects of ablating different aspects of TTCov's optimizer. First, we compare our default approach against Greedy Residual Matching (GRM), which selects samples to minimize the weighted squared error between current AP coverage and the target $B \cdot p^\star$. However, in~\cref{tab:optim_ablation}, GRM yields a dataset with a poorer distributional match to K-Atlas, resulting in lower EPDMS scores. Next, we ablate the parameters within greedy optimization in~\cref{tab:optim_ablation}. We ablate the selection metric by replacing KL divergence with normalized dot product. This swap similarly leads to worse distributional alignment and EPDMS, validating KL divergence as the better objective. While greedy KL selection effectively minimizes global divergence, it can still select samples containing already overly covered, saturated APs. So we add an over coverage penalty $\rho$ that lowers a candidate's score by how much more it adds to saturated APs, with $\rho = 1$ recovering our original score. We observe that both under and over penalizing lead to significantly higher KL divergence and worse EPDMS compared to our baseline, $\rho = 1$. Finally, since Navhard contains hard long-tail scenarios, we employ a standard long-tail strategy, repeat-factor-sampling (RFS)~\cite{gupta2019lvis}, to emphasize rare APs. While RFS naturally increases the KL divergence relative to the unweighted K-Atlas distribution, it reproduces positive long-tail results by significantly improving EPDMS to 26.15, surpassing even the Navtrain oracle. We report our main results without such re-weighting strategies to keep the focus on the core TTCov framework, Atlas. However, these results suggest that further optimization techniques can push TTCov’s performance even higher. For full details on all optimizer ablations, refer to~\cref{app:optim}.

\begin{table}[t]
  \centering
  \caption{Optimizer ablations with corresponding KL divergence to K-Atlas and
  downstream EPDMS.}
  \label{tab:optim_ablation}
  \begingroup
  \small
  \setlength{\tabcolsep}{5pt}
  \renewcommand{\arraystretch}{1.15}
  \begin{tabular*}{\textwidth}{@{\hspace{1.2em}\extracolsep{\fill}}lllcc@{\hspace{1.2em}}}
    \toprule
    Method & Metric & Reweighting & KL div.($\downarrow$) & EPDMS($\uparrow$) \\
    \midrule
    \begin{tabular}[c]{@{}l@{}}Greedy Residual\\Matching\end{tabular}
      & weighted square error & --          & 0.53          & 20.92 \\
    \cdashline{1-5}
    \multirow{7}{*}{\begin{tabular}[c]{@{}l@{}}Greedy Metric\\Optimization\end{tabular}}
      & \multirow{6}{*}{KL divergence}
      & --              & \textbf{0.15} & 24.42 \\
      &                 & RFS, $t=0.001$   & 0.16          & 23.90 \\
      &                 & RFS, $t=0.01$    & 0.18          & 25.39 \\
      &                 & RFS, $t=0.1$     & 0.21          & \textbf{26.15} \\
      &                 & penalty, $\rho=0.5$ & 1.19          & 16.55 \\
      &                 & penalty, $\rho=2.0$ & 0.53          & 21.82 \\
      & Dot product norm
      & --              & 0.45          & 23.37 \\
    \bottomrule
  \end{tabular*}
  \endgroup
  \vspace{-3mm}
\end{table}

\para{Atlas Ablations} We refer to \cref{sec:coverage} and \cref{app:atlas} for Atlas construction ablations and further ablations.

\vspace{-3mm}

\begin{figure}
    \centering
    \includegraphics[width=1.0\linewidth]{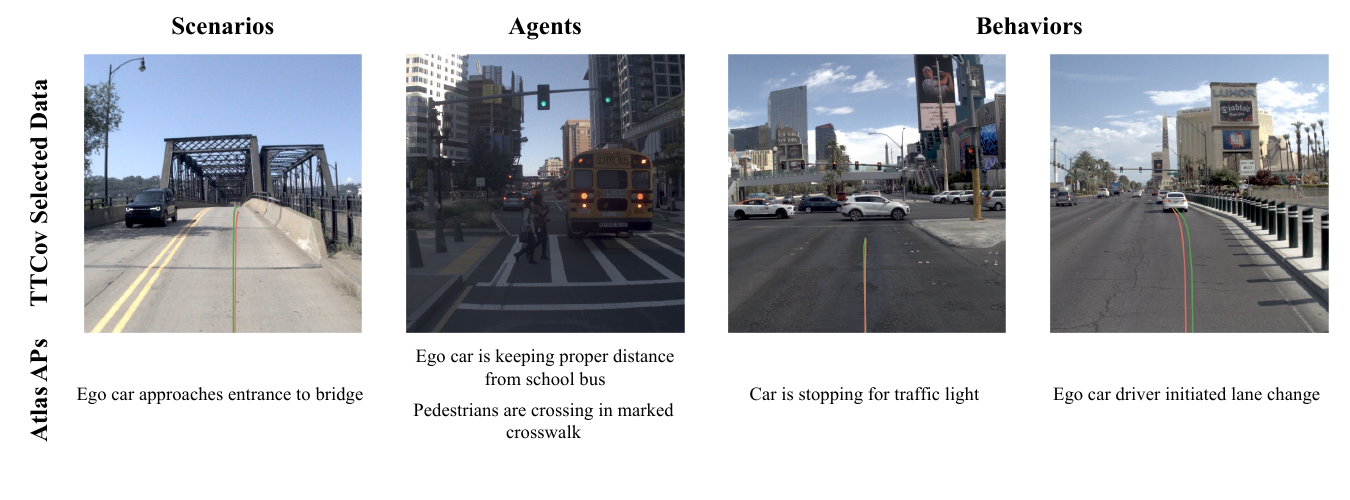}
    \caption{Visual samples of TTCov's selected data with diverse relevant APs from our K-Atlas, demonstrating high-level, low-level, and safety-aware understanding.}
    \label{fig:viz}
    \vspace{-3.5mm}
\end{figure}

\para{Visualizations}
In~\cref{fig:viz}, we illustrate the semantic breadth and depth of TTCov's curation, which selects data whose APs span across a variety of scenarios, agents, and behaviors. Not only do we capture high-level environmental features such as bridges, TTCov also identifies safety critical and usually rare agents like pedestrians and school buses. Finally, TTCov also captures non-nominal driving actions, ensuring the final dataset is both visually and behaviorally diverse. This demonstrates TTCov’s ability beyond curation and towards safety-aware understanding of driving task.

\vspace{-2mm}
\section{Conclusion}
\vspace{-1mm}
We propose TTCov, a framework for test-conditioned curation that adapts training datasets to deployment distributions by leveraging an interpretable Atlas of LLM-based atomic propositions. Unlike traditional curation methods that rely on training-side proxies, TTCov directly optimizes for deployment coverage and distributional alignment. Our results demonstrate that this stronger distributional matching correlates with improved E2E AD performance. Furthermore, while outside of our primary focus, with further optimization improvements, TTCov can surpass even the human-curated oracle, Navtrain. Beyond performance, TTCov demonstrates its scalability and adaptability to novel conditions, common failure points in AD. Through city-to-city experiments, we illustrate the framework's unique ability to adapt to new conditions by curating data for new cities without requiring complete re-curation or suffering from catastrophic forgetting. Lastly, the flexibility of TTCov and its underlying interpretable Atlas demonstrate that the framework can not only meaningfully curate training data, but also capture the nuanced, safety-critical aspects of the driving task.

\vspace{-1.5mm}
\section{Limitations}
\label{sec:lim}
\vspace{-1mm}
While TTCov presents a powerful framework for extracting and leveraging test-conditioned knowledge for data curation, we note two limitations. TTCov efficacy is purposely tied to an existing test set. Thus, TTCov cannot capture information for extremely rare cases that do not exist in the test set nor in open-world knowledge. Similarly, open-world knowledge is extracted via LLMs, limiting TTCov's ceiling to their knowledge. However, LLMs continue to make enormous progress, and TTCov's performance will continue to scale alongside these models.



\bibliographystyle{plain} 
\bibliography{main}

@inproceedings{mo2025kggen,
  title={{KGGen}: Extracting Knowledge Graphs from Plain Text with Language Models},
  author={Mo, Belinda and Yu, Kyssen and Kazdan, Joshua and Mpala, Proud and Yu, Lisa and Kanatsoulis, Charilaos I. and Koyejo, Sanmi},
  booktitle={{NeurIPS}},
  year={2025},
  url={https://openreview.net/forum?id=YyhRJXxbpi}
}

@article{achiam2023gpt,
  title={Gpt-4 technical report},
  author={Achiam, Josh and Adler, Steven and Agarwal, Sandhini and Ahmad, Lama and Akkaya, Ilge and Aleman, Florencia Leoni and Almeida, Diogo and Altenschmidt, Janko and Altman, Sam and Anadkat, Shyamal and others},
  journal={arXiv preprint arXiv:2303.08774},
  year={2023}
}

@article{yang2025qwen3,
  title={Qwen3 technical report},
  author={Yang, An and Li, Anfeng and Yang, Baosong and Zhang, Beichen and Hui, Binyuan and Zheng, Bo and Yu, Bowen and Gao, Chang and Huang, Chengen and Lv, Chenxu and others},
  journal={arXiv preprint arXiv:2505.09388},
  year={2025}
}

@article{comanici2025gemini,
  title={Gemini 2.5: Pushing the frontier with advanced reasoning, multimodality, long context, and next generation agentic capabilities},
  author={Comanici, Gheorghe and Bieber, Eric and Schaekermann, Mike and Pasupat, Ice and Sachdeva, Noveen and Dhillon, Inderjit and Blistein, Marcel and Ram, Ori and Zhang, Dan and Rosen, Evan and others},
  journal={arXiv preprint arXiv:2507.06261},
  year={2025}
}

@inproceedings{gupta2019lvis,
  title={Lvis: A dataset for large vocabulary instance segmentation},
  author={Gupta, Agrim and Dollar, Piotr and Girshick, Ross},
  booktitle={CVPR},
  pages={5356--5364},
  year={2019}
}

@article{ash2019deep,
  title={Deep batch active learning by diverse, uncertain gradient lower bounds},
  author={Ash, Jordan T and Zhang, Chicheng and Krishnamurthy, Akshay and Langford, John and Agarwal, Alekh},
  journal={arXiv preprint arXiv:1906.03671},
  year={2019}
}

@article{abbas2023semdedup,
  title={Semdedup: Data-efficient learning at web-scale through semantic deduplication},
  author={Abbas, Amro and Tirumala, Kushal and Simig, D{\'a}niel and Ganguli, Surya and Morcos, Ari S},
  journal={arXiv preprint arXiv:2303.09540},
  year={2023}
}

@inproceedings{schuhmann2022laion,
  title={{LAION-5B}: An Open Large-Scale Dataset for Training Next Generation Image-Text Models},
  author={Schuhmann, Christoph and Beaumont, Romain and Vencu, Richard and Gordon, Cade and Wightman, Ross and Cherti, Mehdi and Coombes, Theo and Katta, Aarush and Mullis, Clayton and Wortsman, Mitchell and Schramowski, Patrick and Kundurthy, Srivatsa and Crowson, Katherine and Schmidt, Ludwig and Kaczmarczyk, Robert and Jitsev, Jenia},
  booktitle={{NeurIPS}},
  volume={35},
  pages={25278--25294},
  year={2022},
  doi={10.52202/068431-1833}
}

@article{kang2025adadedup,
  title={AdaDeDup: Adaptive Hybrid Data Pruning for Efficient Large-Scale Object Detection Training},
  author={Kang, Feiyang and Chang, Nadine and Shen, Maying and Law, Marc T and Mahmood, Rafid and Jia, Ruoxi and Alvarez, Jose M},
  journal={arXiv preprint arXiv:2507.00049},
  year={2025}
}

@inproceedings{slyman2024fairdedup,
  title={Fairdedup: Detecting and mitigating vision-language fairness disparities in semantic dataset deduplication},
  author={Slyman, Eric and Lee, Stefan and Cohen, Scott and Kafle, Kushal},
  booktitle={CVPR},
  pages={13905--13916},
  year={2024}
}

@article{ren2021survey,
  title={A Survey of Deep Active Learning},
  author={Ren, Pengzhen and Xiao, Yun and Chang, Xiaojun and Huang, Po-Yao and Li, Zhihui and Gupta, Brij B. and Chen, Xiaojiang and Wang, Xin},
  journal={{CSUR}},
  volume={54},
  number={9},
  pages={180:1--180:40},
  year={2021},
  doi={10.1145/3472291},
  publisher={ACM}
}

@inproceedings{joshi2009multi,
  title={Multi-class active learning for image classification},
  author={Joshi, Ajay J and Porikli, Fatih and Papanikolopoulos, Nikolaos},
  booktitle={CVPR},
  pages={2372--2379},
  year={2009},
  organization={IEEE}
}

@inproceedings{paul2021deep,
  title={Deep Learning on a Data Diet: Finding Important Examples Early in Training},
  author={Paul, Mansheej and Ganguli, Surya and Dziugaite, Gintare Karolina},
  booktitle={{NeurIPS}},
  volume={34},
  pages={20596--20607},
  year={2021}
}

@inproceedings{
toneva2018an,
title={An Empirical Study of Example Forgetting during Deep Neural Network Learning},
author={Mariya Toneva and Alessandro Sordoni and Remi Tachet des Combes and Adam Trischler and Yoshua Bengio and Geoffrey J. Gordon},
booktitle={{ICLR}},
year={2019},
url={https://openreview.net/forum?id=BJlxm30cKm},
}

@inproceedings{codevilla2018end,
  title={End-to-end driving via conditional imitation learning},
  author={Codevilla, Felipe and M{\"u}ller, Matthias and L{\'o}pez, Antonio and Koltun, Vladlen and Dosovitskiy, Alexey},
  booktitle={ICRA},
  pages={4693--4700},
  year={2018},
  organization={IEEE}
}

@article{chitta2023transfuser,
  title={{TransFuser}: Imitation with Transformer-Based Sensor Fusion for Autonomous Driving},
  author={Chitta, Kashyap and Prakash, Aditya and Jaeger, Bernhard and Yu, Zehao and Renz, Katrin and Geiger, Andreas},
  journal={{TPAMI}},
  volume={45},
  number={11},
  pages={12878--12895},
  year={2023},
  doi={10.1109/TPAMI.2022.3200245},
  publisher={IEEE}
}

@InProceedings{jiang2023vad,
    author    = {Jiang, Bo and Chen, Shaoyu and Xu, Qing and Liao, Bencheng and Chen, Jiajie and Zhou, Helong and Zhang, Qian and Liu, Wenyu and Huang, Chang and Wang, Xinggang},
    title     = {VAD: Vectorized Scene Representation for Efficient Autonomous Driving},
    booktitle = {ICCV},
    month     = {October},
    year      = {2023},
    pages     = {8340-8350}
}

@article{chen2024vadv2,
  title={Vadv2: End-to-end vectorized autonomous driving via probabilistic planning},
  author={Chen, Shaoyu and Jiang, Bo and Gao, Hao and Liao, Bencheng and Xu, Qing and Zhang, Qian and Huang, Chang and Liu, Wenyu and Wang, Xinggang},
  journal={arXiv preprint arXiv:2402.13243},
  year={2024}
}

@article{li2024hydra,
  title={Hydra-mdp: End-to-end multimodal planning with multi-target hydra-distillation},
  author={Li, Zhenxin and Li, Kailin and Wang, Shihao and Lan, Shiyi and Yu, Zhiding and Ji, Yishen and Li, Zhiqi and Zhu, Ziyue and Kautz, Jan and Wu, Zuxuan and others},
  journal={arXiv preprint arXiv:2406.06978},
  year={2024}
}

@inproceedings{li2024ego,
  title={Is ego status all you need for open-loop end-to-end autonomous driving?},
  author={Li, Zhiqi and Yu, Zhiding and Lan, Shiyi and Li, Jiahan and Kautz, Jan and Lu, Tong and Alvarez, Jose M},
  booktitle={CVPR},
  pages={14864--14873},
  year={2024}
}

@inproceedings{dauner2023parting,
  author       = {Daniel Dauner and Marcel Hallgarten and Andreas Geiger and Kashyap Chitta},
  title        = {Parting with Misconceptions about Learning-based Vehicle Motion Planning},
  year         = {2023},
  booktitle    = {CoRL},
}

@inproceedings{goyal2024scaling,
  title={Scaling Laws for Data Filtering--Data Curation cannot be Compute Agnostic},
  author={Goyal, Sachin and Maini, Pratyush and Lipton, Zachary C and Raghunathan, Aditi and Kolter, J Zico},
  booktitle={CVPR},
  pages={22702--22711},
  year={2024}
}

@inproceedings{sorscher2022beyond,
  title={Beyond Neural Scaling Laws: Beating Power Law Scaling via Data Pruning},
  author={Sorscher, Ben and Geirhos, Robert and Shekhar, Shashank and Ganguli, Surya and Morcos, Ari S.},
  booktitle={{NeurIPS}},
  volume={35},
  pages={19523--19536},
  year={2022},
  doi={10.52202/068431-1419}
}

@incollection{lewis1994heterogeneous,
  title={Heterogeneous Uncertainty Sampling for Supervised Learning},
  author={Lewis, David D. and Catlett, Jason},
  booktitle={{ICML}},
  pages={148--156},
  year={1994},
  publisher={Elsevier}
}

@inproceedings{smith2023prediction,
  title={Prediction-oriented bayesian active learning},
  author={Smith, Freddie Bickford and Kirsch, Andreas and Farquhar, Sebastian and Gal, Yarin and Foster, Adam and Rainforth, Tom},
  booktitle={AISTATS},
  pages={7331--7348},
  year={2023},
}

@inproceedings{dimlioglu2026scaling,
  title={Scaling-Aware Data Selection for End-to-End Autonomous Driving Systems},
  author={Dimlioglu, Tolga and Chang, Nadine and Shen, Maying and Mahmood, Rafid and Alvarez, Jose M.},
  booktitle={{CVPR}},
  year={2026},
  note={Accepted}
}

@article{li2025ztrs,
  title={Ztrs: Zero-imitation end-to-end autonomous driving with trajectory scoring},
  author={Li, Zhenxin and Yao, Wenhao and Wang, Zi and Sun, Xinglong and Chen, Jingde and Chang, Nadine and Shen, Maying and Song, Jingyu and Wu, Zuxuan and Lan, Shiyi and others},
  journal={arXiv preprint arXiv:2510.24108},
  year={2025}
}

@inproceedings{joachims1999transductive,
    title={Transductive Inference for Text Classification using Support Vector Machines},
    author={Joachims, Thorsten},
    booktitle={ICML},
    year={1999}
}

@inproceedings{bennett1999semi,
    title={Semi-Supervised Support Vector Machines},
    author={Bennett, Kristin P. and Demiriz, Ayhan},
    booktitle={NeurIPS},
    year={1998}
  }

@inproceedings{zhu2003semi,
    title={Semi-Supervised Learning Using {Gaussian} Fields and Harmonic Functions},
    author={Zhu, Xiaojin and Ghahramani, Zoubin and Lafferty, John},
    booktitle={ICML},
    year={2003}
  }

@inproceedings{zhou2004learning,
    title={Learning with Local and Global Consistency},
    author={Zhou, Dengyong and Bousquet, Olivier and Lal, Thomas and Weston, Jason and Sch{\"o}lkopf, Bernhard},
    booktitle={NeurIPS},
    year={2004}
  }

@article{belkin2006manifold,
  title={Manifold Regularization: A Geometric Framework for Learning from Labeled and Unlabeled Examples},
  author={Belkin, Mikhail and Niyogi, Partha and Sindhwani, Vikas},
  journal={{JMLR}},
  volume={7},
  pages={2399--2434},
  year={2006}
}

@book{vapnik2013nature,
  title={The nature of statistical learning theory},
  author={Vapnik, Vladimir},
  year={2013},
  publisher={Springer science \& business media}
}

@article{cao2025pseudo,
  title={Pseudo-simulation for autonomous driving},
  author={Cao, Wei and Hallgarten, Marcel and Li, Tianyu and Dauner, Daniel and Gu, Xunjiang and Wang, Caojun and Miron, Yakov and Aiello, Marco and Li, Hongyang and Gilitschenski, Igor and others},
  journal={arXiv preprint arXiv:2506.04218},
  year={2025}
}

@inproceedings{hardt2024test,
  title={Test-Time Training on Nearest Neighbors for Large Language Models},
  author={Hardt, Moritz and Sun, Yu},
  booktitle={ICLR},
  year={2024}
}

@inproceedings{niu2023sar,
    title={Towards Stable Test-Time Adaptation in Dynamic Wild World},
    author={Niu, Shuaicheng and Wu, Jiaxiang and Zhang, Yifan and Wen, Zhiquan and Chen, Yaofo and Zhao, Peilin and Tan, Mingkui},
    booktitle={ICLR},
    year={2023}
  }

@inproceedings{zhang2022memo,
    title={{MEMO}: Test Time Robustness via Adaptation and Augmentation},
    author={Zhang, Marvin and Levine, Sergey and Finn, Chelsea},
    booktitle={NeurIPS},
    year={2022}
  }

@inproceedings{sun2020ttt,
  title={Test-time training with self-supervision for generalization under distribution shifts},
  author={Sun, Yu and Wang, Xiaolong and Liu, Zhuang and Miller, John and Efros, Alexei and Hardt, Moritz},
  booktitle={ICML},
  pages={9229--9248},
  year={2020},
  organization={PMLR}
}

@article{sima2025centaur,
  title={Centaur: Robust end-to-end autonomous driving with test-time training},
  author={Sima, Chonghao and Chitta, Kashyap and Yu, Zhiding and Lan, Shiyi and Luo, Ping and Geiger, Andreas and Li, Hongyang and Alvarez, Jose M},
  journal={arXiv preprint arXiv:2503.11650},
  year={2025}
}

@inproceedings{wang2021tent,
  title={{Tent}: Fully Test-Time Adaptation by Entropy Minimization},
  author={Wang, Dequan and Shelhamer, Evan and Liu, Shaoteng and Olshausen, Bruno and Darrell, Trevor},
  booktitle={{ICLR}},
  year={2021},
  url={https://openreview.net/forum?id=uXl3bZLkr3c}
}

@inproceedings{sener2018coreset,
  title     = {Active Learning for Convolutional Neural Networks: A Core-Set Approach},
  author    = {Sener, Ozan and Savarese, Silvio},
  booktitle = {ICLR},
  year      = {2018},
  url={https://openreview.net/forum?id=H1aIuk-RW}
}

@inproceedings{coleman2020selection,
  title={Selection via Proxy: Efficient Data Selection for Deep Learning},
  author={Coleman, Cody and Yeh, Christopher and Mussmann, Stephen and Mirzasoleiman, Baharan and Bailis, Peter and Liang, Percy and Leskovec, Jure and Zaharia, Matei},
  booktitle={{ICLR}},
  year={2020},
  url={https://openreview.net/forum?id=HJg2b0VYDr}
}

@inproceedings{shen2025sse,
  title={Sse: Multimodal semantic data selection and enrichment for industrial-scale data assimilation},
  author={Shen, Maying and Chang, Nadine and Liu, Sifei and Alvarez, Jose M},
  booktitle={KDD},
  pages={2525--2535},
  year={2025}
}

@inproceedings{clip2021,
  title     = {Learning Transferable Visual Models From Natural Language Supervision},
  author    = {Radford, Alec and Kim, Jong Wook and Hallacy, Chris and Ramesh, Aditya and Goh, Gabriel and Agarwal, Sandhini and Sastry, Girish and Askell, Amanda and Mishkin, Pamela and Clark, Jack and Krueger, Gretchen and Sutskever, Ilya},
  booktitle = {ICML},
  year      = {2021},
}

@inproceedings{liu2021ttt++,
  title={{TTT++}: When Does Self-Supervised Test-Time Training Fail or Thrive?},
  author={Liu, Yuejiang and Kothari, Parth and van Delft, Bastien and Bellot-Gurlet, Baptiste and Mordan, Taylor and Alahi, Alexandre},
  booktitle={{NeurIPS}},
  volume={34},
  pages={21808--21820},
  year={2021}
}

@inproceedings{caesar2020nuscenes,
  title={{nuScenes}: A Multimodal Dataset for Autonomous Driving},
  author={Caesar, Holger and Bankiti, Varun and Lang, Alex H. and Vora, Sourabh and Liong, Venice Erin and Xu, Qiang and Krishnan, Anush and Pan, Yu and Baldan, Giancarlo and Beijbom, Oscar},
  booktitle={{CVPR}},
  pages={11621--11631},
  year={2020},
  doi={10.1109/CVPR42600.2020.01164}
}

@inproceedings{sun2020waymo,
  title={Scalability in Perception for Autonomous Driving: Waymo Open Dataset},
  author={Sun, Pei and Kretzschmar, Henrik and Dotiwalla, Xerxes and Chouard, Aurelien and Patnaik, Vijaysai and Tsui, Paul and Guo, James and Zhou, Yin and Chai, Yuning and Caine, Benjamin and Vasudevan, Vijay and Han, Wei and Ngiam, Jiquan and Zhao, Hang and Timofeev, Aleksei and Ettinger, Scott and Krivokon, Maxim and Gao, Amy and Joshi, Aditya and Zhang, Yu and Shlens, Jonathon and Chen, Zhifeng and Anguelov, Dragomir},
  booktitle={{CVPR}},
  pages={2446--2454},
  year={2020},
  doi={10.1109/CVPR42600.2020.00252}
}

@article{liao2022kitti360,
  title={{KITTI-360}: A Novel Dataset and Benchmarks for Urban Scene Understanding in 2D and 3D},
  author={Liao, Yiyi and Xie, Jun and Geiger, Andreas},
  journal={{TPAMI}},
  volume={45},
  number={3},
  pages={3292--3310},
  year={2023},
  doi={10.1109/TPAMI.2022.3179507}
}

@article{pattnayak2024survey,
  title={Survey of Large Multimodal Model Datasets, Application Categories and Taxonomy},
  author={Pattnayak, Priyaranjan and Patel, Hitesh Laxmichand and Kumar, Bhargava and Agarwal, Amit and Banerjee, Ishan and Panda, Srikant and Kumar, Tejaswini},
  journal={arXiv preprint arXiv:2412.17759},
  year={2024}
}

@article{liu2024survey,
  title={A Survey on Autonomous Driving Datasets: Statistics, Annotation Quality, and a Future Outlook},
  author={Liu, Mingyu and Yurtsever, Ekim and Fossaert, Jonathan and Zhou, Xingcheng and Zimmer, Walter and Cui, Yuning and Zagar, Bare Luka and Knoll, Alois C.},
  journal={{TIV}},
  volume={9},
  number={11},
  pages={7138--7164},
  year={2024},
  doi={10.1109/TIV.2024.3394735}
}

@misc{Grzywaczewski2017NvidiaScale,
  title={Training {AI} for Self-Driving Vehicles: The Challenge of Scale},
  author={Grzywaczewski, Adam},
  howpublished={{NVIDIA Technical Blog}},
  year={2017},
  month={Oct.},
  url={https://developer.nvidia.com/blog/training-self-driving-vehicles-challenge-scale/},
  note={Accessed: 2026-05-06}
}

@misc{Coren2025TeslaData,
  title={{Tesla} Has 780 Million Miles of Driving Data, and Adds Another Million Every 10 Hours},
  author={Coren, Michael J.},
  howpublished={{Quartz}},
  year={2016},
  url={https://qz.com/694520/tesla-has-780-million-miles-of-driving-data-and-adds-another-million-every-10-hours},
  note={Published May 28, 2016. Accessed: 2026-05-06}
}

@inproceedings{chang2026position,
  title     = {Position: Stop Reactively Patching Your Model Every Time and Start Proactive Test-Driven AI Development},
  author    = {Chang, Nadine and Shen, Maying and Wang, Jialiang and Mahmood, Rafid and Alvarez, Jose M.},
  booktitle = {ICML},
  year      = {2026}
}

@inproceedings{wang2022cotta,
  title={Continual Test-Time Domain Adaptation},
  author={Wang, Qin and Fink, Olga and Van Gool, Luc and Dai, Dengxin},
  booktitle={{CVPR}},
  pages={7201--7211},
  year={2022},
  doi={10.1109/CVPR52688.2022.00706}
}

@inproceedings{niu2022efficient,
  title={Efficient Test-Time Model Adaptation without Forgetting},
  author={Niu, Shuaicheng and Wu, Jiaxiang and Zhang, Yifan and Chen, Yaofo and Zheng, Shijian and Zhao, Peilin and Tan, Mingkui},
  booktitle={{ICML}},
  volume={162},
  pages={16888--16905},
  year={2022},
  url={https://proceedings.mlr.press/v162/niu22a.html}
}

@article{douze2025faiss,
  title={The {Faiss} Library},
  author={Douze, Matthijs and Guzhva, Alexandr and Deng, Chengqi and Johnson, Jeff and Szilvasy, Gergely and Mazar{\'e}, Pierre-Emmanuel and Lomeli, Maria and Hosseini, Lucas and J{\'e}gou, Herv{\'e}},
  journal={{TBD}},
  pages={1--17},
  year={2025},
  doi={10.1109/TBDATA.2025.3618474},
  publisher={IEEE}
}

@inproceedings{dauner2024navsim,
  title={{NAVSIM}: Data-Driven Non-Reactive Autonomous Vehicle Simulation and Benchmarking},
  author={Dauner, Daniel and Hallgarten, Marcel and Li, Tianyu and Weng, Xinshuo and Huang, Zhiyu and Yang, Zetong and Li, Hongyang and Gilitschenski, Igor and Ivanovic, Boris and Pavone, Marco and Geiger, Andreas and Chitta, Kashyap},
  booktitle={NeurIPS},
  volume={37},
  year={2024}
}

@misc{openscene2023,
  title={OpenScene: The Largest Up-to-Date 3D Occupancy Prediction Benchmark in Autonomous Driving},
  author={OpenScene Contributors},
  howpublished={\url{https://github.com/OpenDriveLab/OpenScene}},
  year={2023}
}

\newpage
\appendix
\section{Broader Impact}
\label{app:broader}
TTCov improves autonomous driving safety by enabling interpretable, deployment-conditioned data curation. By guiding selection in an explicit Atlas of atomic propositions, TTCov enables adaptability to novel environments with potentially new edge-cases and reduces the computational cost for learning to new domains. However, we note that real-world deployment still requires rigorous closed-loop testing and oversight.

\section{Atlas Ablations}
\label{app:atlas}

\subsection{Atlas threshold for AP matching}
To additionally measure the TTCov's robustness to LLM sensitivities, we ablate the cosine similarity threshold used by Qwen 3 in AP matching to cluster similar APs. Within each cluster, an LLM verifies all phrases are semantically similar, finds any outliers, and merges all phrases into a single phrase. As seen in~\cref{tab:thres}, TTCov is robust to this threshold and EPDMS scores remain consistent across all thresholds. 

\begin{table}[h]
    \centering %
    
    \caption{We ablate the threshold used by Qwen 3 in AP matching to cluster similar APs. Within each cluster, an LLM (Gemini 2.5 Pro) performs all final matches and merging. Due to compute constraints, reported numbers are from one run.}
    \label{tab:thres}
    \vspace{1mm}
    \begin{tabular}{lcc}
          \toprule
          Threshold & EPDMS \\
          \midrule
          0.83 & 24.12 \\
          0.84 & 24.44 \\
          \textbf{0.85 (main)} & 24.42 \\
          0.86 & 24.54 \\
          0.87 & 24.36 \\
          \bottomrule
    \end{tabular}
\end{table}

\subsection{Why LLM atomic propositions? Continued.}
\begin{figure}[h]
    \centering
    \includegraphics[width=1.0\linewidth]{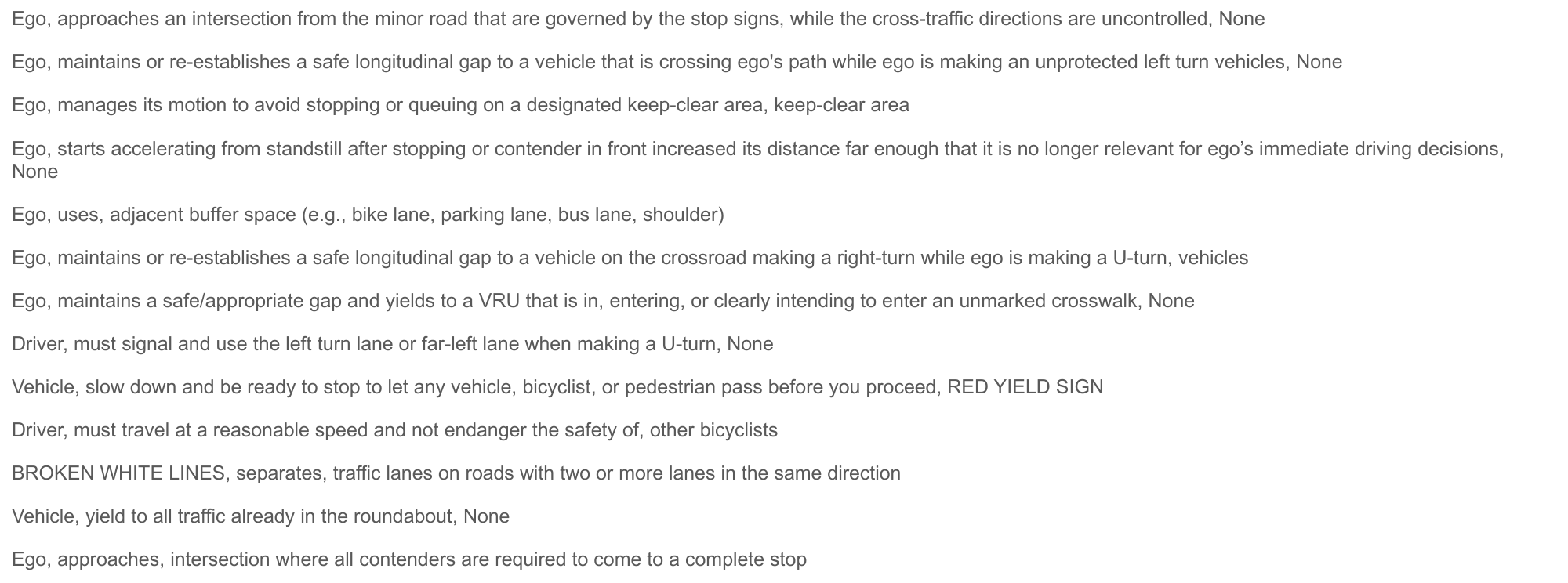}
    \caption{Samples of complex triplets in knowledge graph.}
    \label{fig:kg}

\end{figure}

\begin{figure}[h]
    \centering
    \includegraphics[width=.9\linewidth]{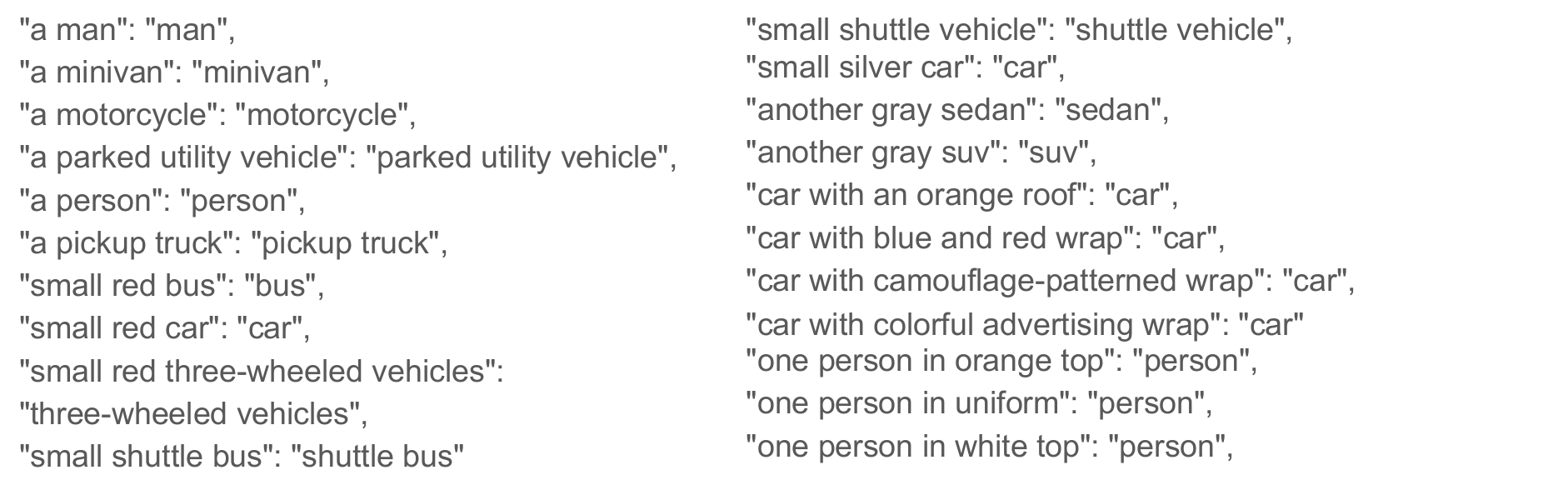}
    \caption{Samples of unnecessary modifiers, which we removed in our final de-noising phrase unification stage.}
    \label{fig:modifiers}

\end{figure}

In~\cref{fig:kg}, we illustrate some complex triplets in a knowledge graph extracted from a large text corpus. We observe that each triplet contains phrases that are difficult to parse and contain a multitude of information. Furthermore, triplets often end with \enquote{None}. Leveraging such a complex knowledge graph representation makes it difficult to find key commonalities among test data. Next, we provide additional samples of unnecessary modifiers that are irrelevant for our task in~\cref{fig:modifiers}. Modifiers include basic articles, colors, and clothing. In our final de-noising and unification stage AP extraction, we merge semantically similar phrases, if not for these modifiers.

\subsection{Structured ontology extracted from LLM}
\begin{figure}[h]
    \centering
    \includegraphics[width=1.0\linewidth]{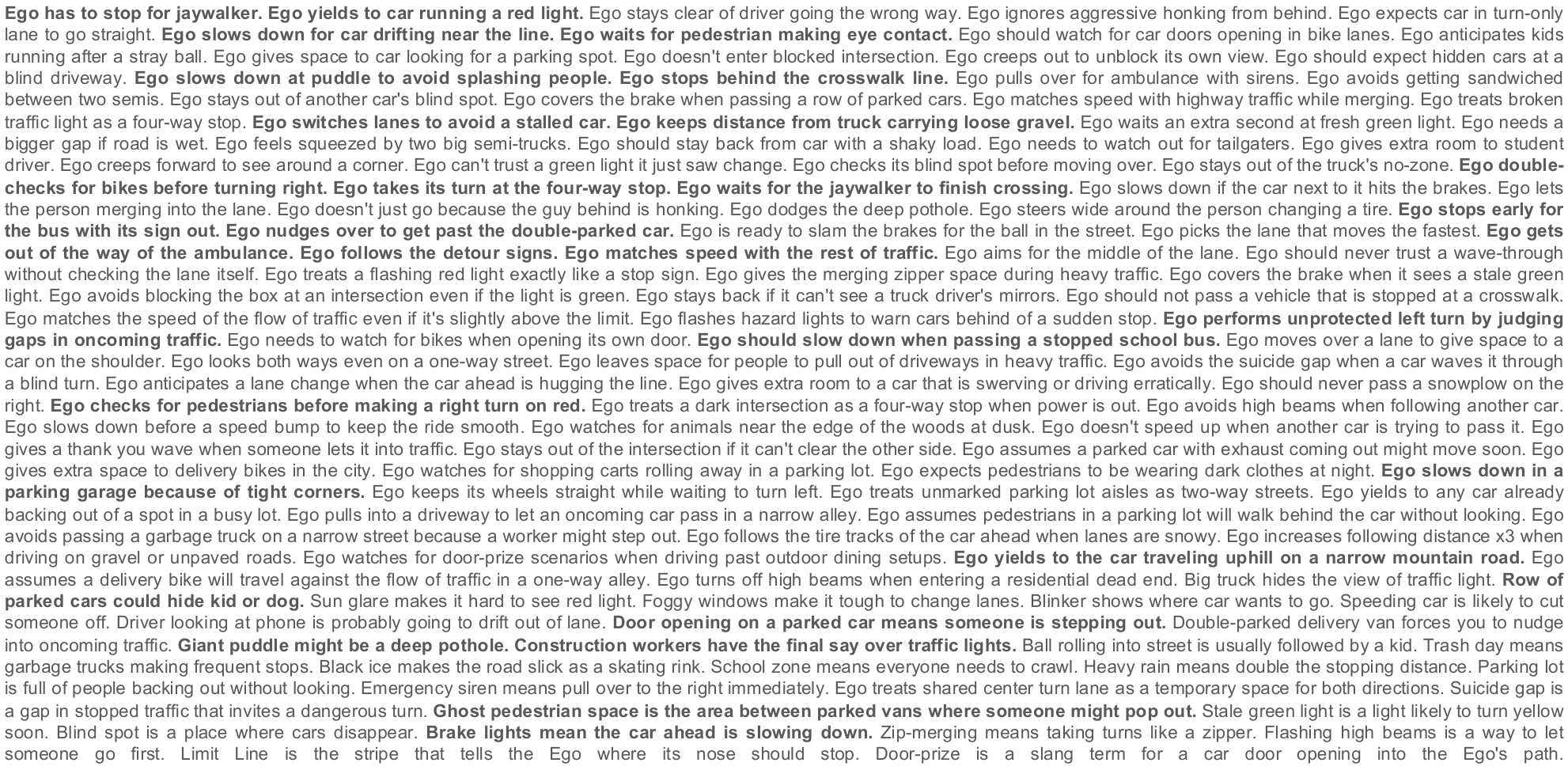}
    \caption{Samples of decomposed APs prior to de-noising and unification from LLM generated ontologies.}
    \label{fig:ontology}

\end{figure}

We prompt Gemini 2.5 Pro, trained on internet-scale data, to generate strong ontologies for autonomous driving. We illustrate a subsample of decomposed APs prior to de-noising and unification from LLM generated ontologies in~\cref{fig:ontology}. Critically, we observe the ontology includes diverse ego actions as well as important driving events to note (\eg \enquote{Ego waits for the jaywalker to finish crossing.} and \enquote{Giant puddle might be a deep pothole.})

\subsection{Additional coverage analysis} 

\begin{figure}[h]
    \centering
    \includegraphics[width=1.0\linewidth]{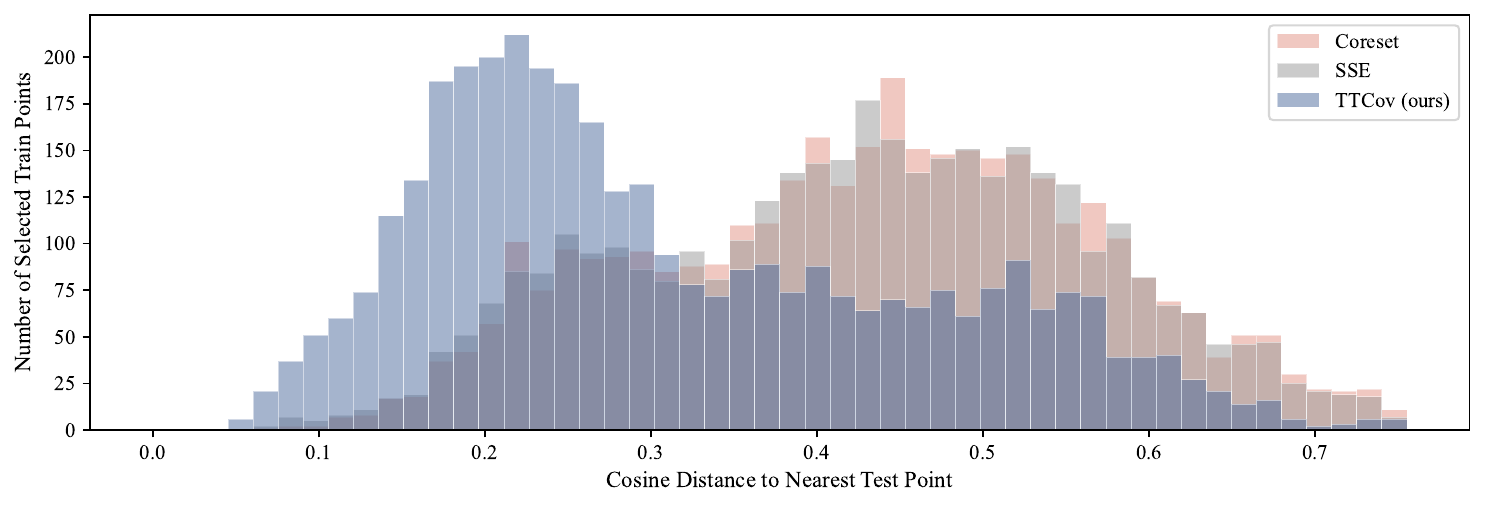}
    \caption{TTCov's selected data coverage over the 2nd test city. We show the histogram of the number of selected data points, whose embeddings are within a certain distance from any nearest test points. Given only one target test city, TTCov selects more data points closer to test points (see left of fig).
    }
    \label{fig:city2_cover}
\end{figure}

\begin{figure}[h]
    \centering
    \includegraphics[width=0.85\linewidth]{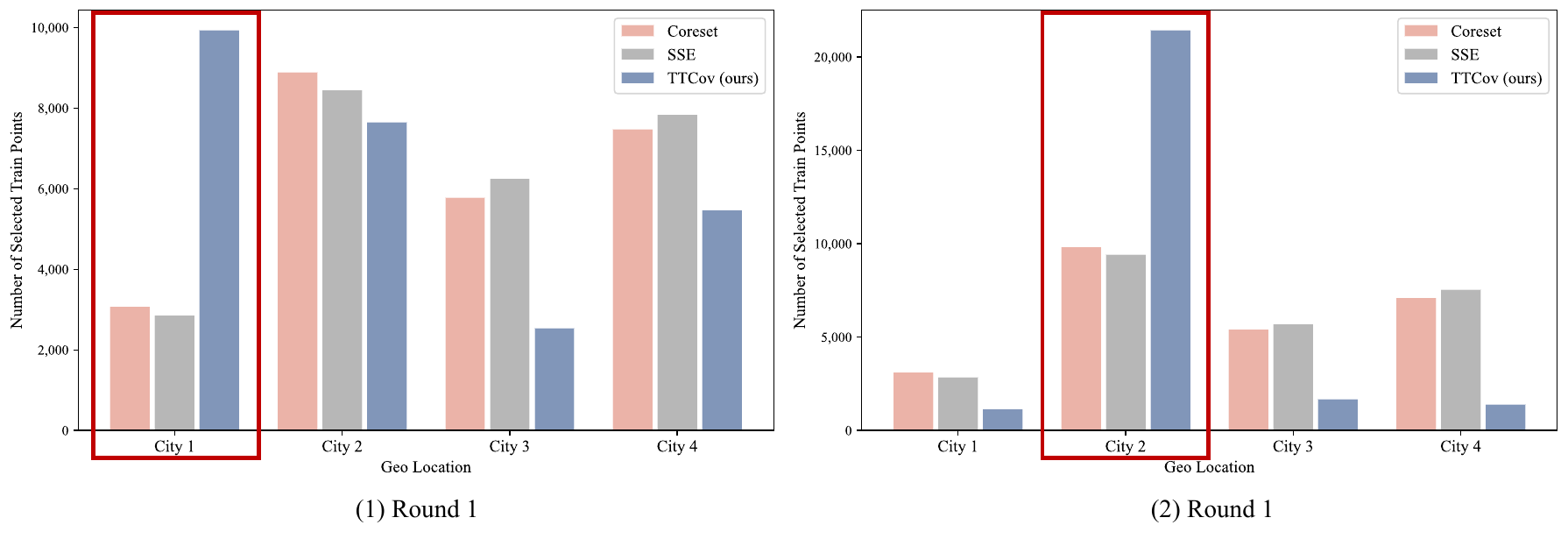}
    \caption{The geography distribution of selected data during the first and second round of city evolution. In each round, TTCov's selected data contains more data points that are from the target test city.
    }
    \label{fig:city_geo}
\end{figure}

We visualize TTCov's  selected data coverage for the second test city in \cref{fig:city2_cover}. Consistent with the patterns observed for City 1, TTCov selects a higher volume of training points in close proximity to the test distribution than baseline methods. Furthermore, \cref{fig:city_geo} illustrates the ground-truth geographic distribution of the data selected by TTCov. The results show that TTCov targets the specific test city significantly better by selecting more than 3$\times$ the data for City 1 and 2$\times$ for City 2 compared to other methods. Additionally, we show more quantitative coverage metrics in~\cref{tab:cover_all_metrics}, where we observe TTCov selects high volume of training points even as we increase the distance to nearest test data. We also report an additional metric, mean NN distance, an average of all distances for each train point to nearest test point. Again, TTCov selects a tighter set of points around test regions. These results further validate TTCov’s ability to maximize relevant data coverage even in dynamically changing test domains.

\begin{table}[h!]
  \centering
  \caption{TTCov's coverage metrics for multi-round city evolution.}
  \label{tab:cover_all_metrics}
  \begingroup
  \small
  \setlength{\tabcolsep}{4pt}
  \renewcommand{\arraystretch}{1.12}
  \begin{tabular*}{\textwidth}{@{\hspace{0.8em}\extracolsep{\fill}}ccccccc@{\hspace{0.8em}}}
    \toprule
    & & \multicolumn{3}{c}{Num. Selected Train($\uparrow$)} & & \\
    \cmidrule(lr){3-5}
    Round & Method & NN@0.15 & NN@0.30 & NN@0.45 & Mean NN Dist.($\downarrow$) & MMD($\downarrow$) \\
    \midrule
    \multirow{3}{*}{Round 1}
      & Coreset       & 42  & 659  & 1646 & 0.20 & 0.13 \\
      & SSE           & 46  & 699  & 1697 & 0.19 & 0.13 \\
      & \textbf{TTCov (ours)}
                      & \textbf{732} & \textbf{1744} & \textbf{2436}
                      & \textbf{0.16} & \textbf{0.08} \\
    \cmidrule(lr){1-7}
    \multirow{3}{*}{Round 2}
      & Coreset       & 36  & 733  & 1950 & 0.19 & 0.10 \\
      & SSE           & 49  & 774  & 1991 & 0.18 & 0.11 \\
      & \textbf{TTCov (ours)}
                      & \textbf{355} & \textbf{2083} & \textbf{2877}
                      & \textbf{0.16} & \textbf{0.06} \\
    \cmidrule(lr){1-7}
    \multirow{3}{*}{Round 3}
      & Coreset       & 19  & 357  & 849  & 0.25 & 0.20 \\
      & SSE           & 17  & 362  & 856  & 0.25 & 0.20 \\
      & \textbf{TTCov (ours)}
                      & \textbf{47} & \textbf{814} & \textbf{1651}
                      & \textbf{0.23} & \textbf{0.14} \\
    \cmidrule(lr){1-7}
    \multirow{3}{*}{Round 4}
      & Coreset       & 62  & 785  & 2019 & 0.19 & 0.13 \\
      & SSE           & 68  & 837  & 2096 & 0.19 & 0.13 \\
      & \textbf{TTCov (ours)}
                      & \textbf{228} & \textbf{1341} & \textbf{2532}
                      & \textbf{0.17} & \textbf{0.10} \\
    \bottomrule
  \end{tabular*}
  \endgroup
\end{table}

\subsection{Prompts}
\begin{figure}[h!]
    \centering
    \begin{subfigure}[b]{1.0\linewidth}
        \centering
        \includegraphics[width=\linewidth]{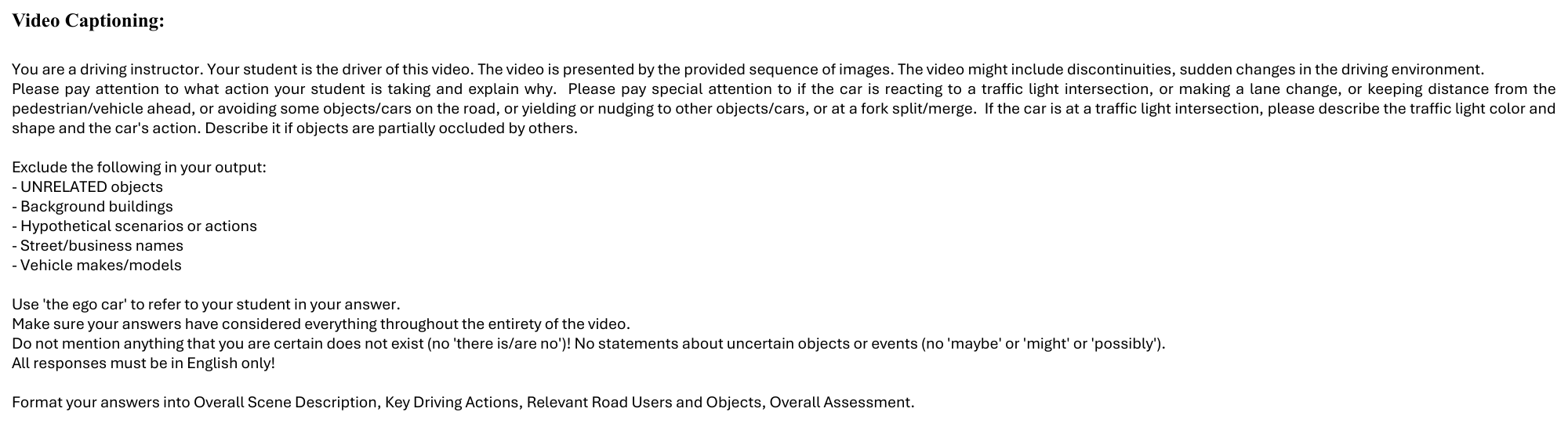}
        \caption{Prompt for video knowledge extraction.}
        \label{fig:prompt_caption}
    \end{subfigure}
    
    \vspace{1em} 

    \begin{subfigure}[b]{1.0\linewidth}
        \centering
        \includegraphics[width=\linewidth]{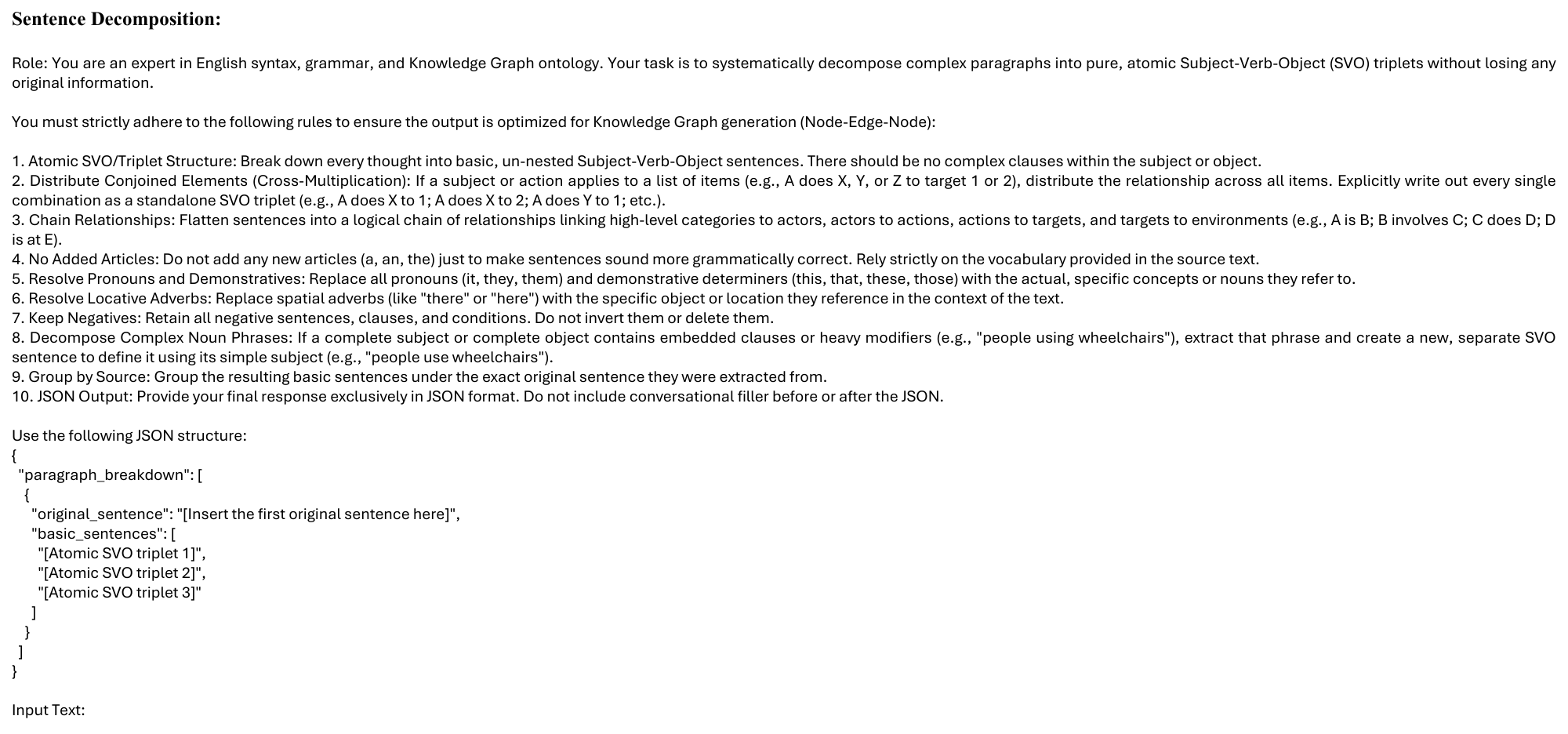}
        \caption{Prompt for decomposing raw knowledge to atomic propositions, APs extraction.}
        \label{fig:prompt_decompose}
    \end{subfigure}

    \begin{subfigure}[b]{1.0\linewidth}
        \centering
        \includegraphics[width=\linewidth]{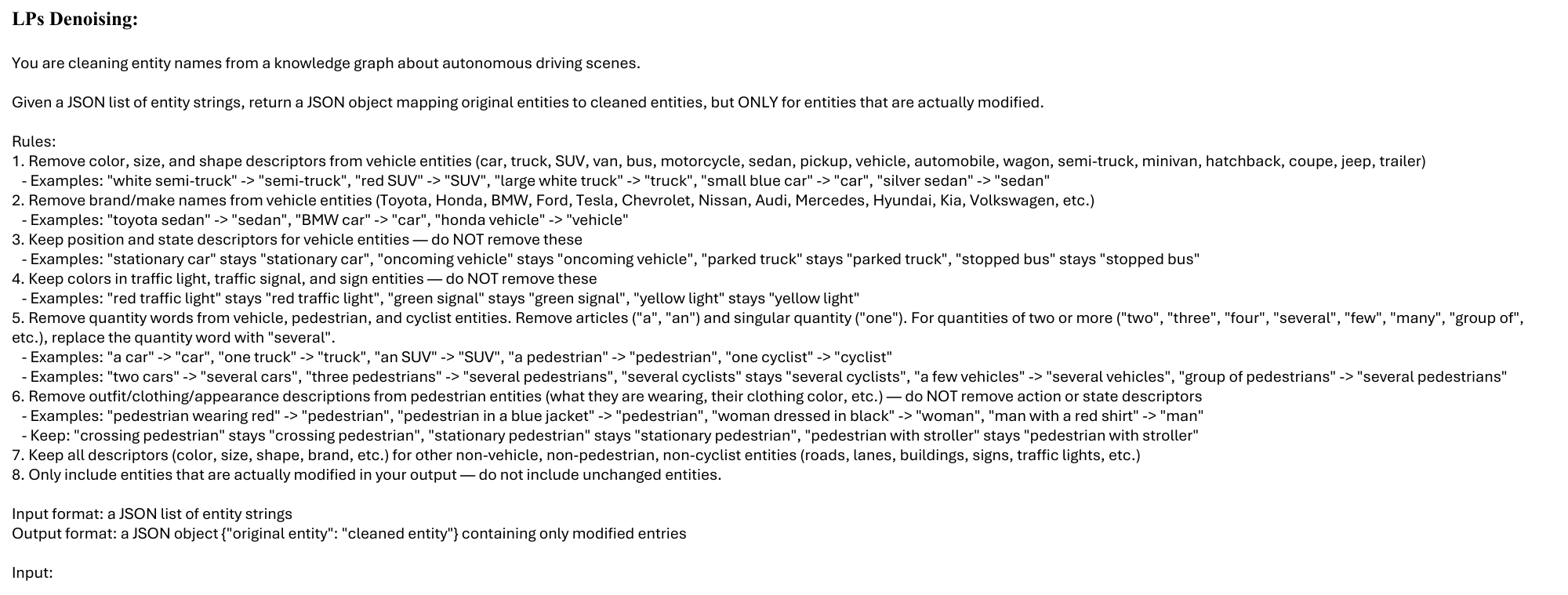}
        \caption{Prompt for APs de-noising.}
        \label{fig:lp_denoise}
    \end{subfigure}

    \begin{subfigure}[b]{1.0\linewidth}
        \centering
        \includegraphics[width=\linewidth]{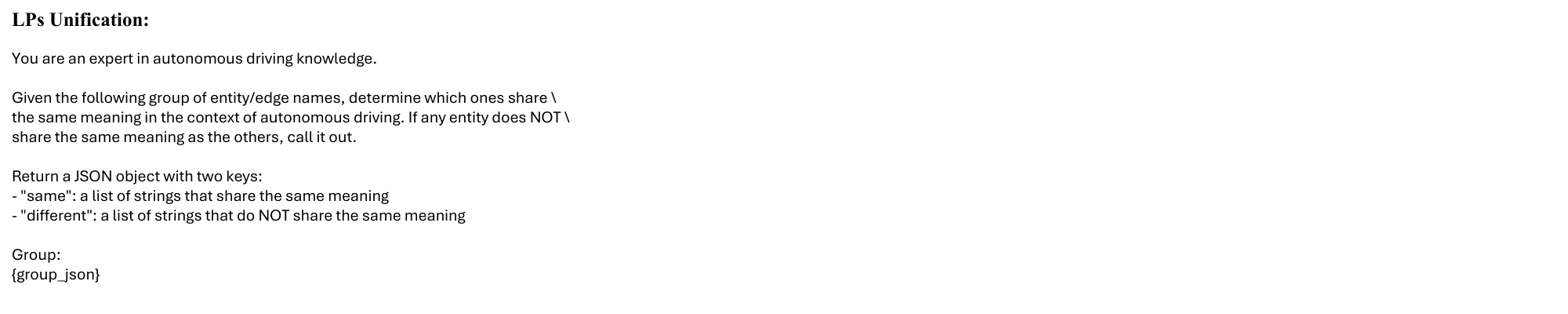}
        \caption{Prompt for APs unification.}
        \label{fig:lp_unification}
    \end{subfigure}
    
    \caption{All relevant prompts used throughout TTCov's process.}
    \label{fig:prompts}
\end{figure}

We prompt Gemini 2.5 Pro for video knowledge extraction, APs extraction, APs de-noising and APs unification. We provide the prompts in ~\cref{fig:prompts}.

\section{EPDMS Submetrics}
\label{app:metrics}
\para{Additional Results}
In~\cref{tab:full_metrics}, we report the full set of submetrics corresponding to~\cref{tab:main_tab}. 

\begin{table*}[!]
  \centering
  \caption{Additional subscores across all curation experiments corresponding to~\cref{tab:main_tab}.
  Each cell reports the average score with standard deviation shown as subscript. Same overall EPDMS across Stage 1 and 2 reported in last column.}
  \label{tab:full_metrics}
  \begingroup
  \footnotesize
  \setlength{\tabcolsep}{3pt}
  \renewcommand{\arraystretch}{1.12}
  \newcommand{\avgstd}[2]{\ensuremath{#1_{\scriptscriptstyle \pm #2}}}
  \textbf{Stage 1}

  \vspace{0.2em}

  \resizebox{\textwidth}{!}{%
  \begin{tabular}{@{\hspace{0.4em}}cccccccccccc@{\hspace{0.4em}}}
    \toprule
    Budget & Method & NC & DAC & DDC & TLC & EP & TTC & LK & HC & EC & EPDMS \\
    \midrule
    \multirow{4}{*}{0.5x} & Random & \avgstd{92.74}{0.96} & \avgstd{63.11}{1.60} & \avgstd{91.81}{0.94} & \avgstd{99.56}{0.22} & \avgstd{80.86}{0.54} & \avgstd{91.33}{1.18} & \avgstd{86.07}{0.68} & \avgstd{97.70}{0.13} & \avgstd{77.48}{0.68} & \avgstd{18.95}{0.39} \\
    & Coreset & \avgstd{94.74}{0.55} & \avgstd{69.56}{1.39} & \avgstd{95.78}{0.84} & \avgstd{99.33}{0.00} & \avgstd{81.97}{0.75} & \avgstd{92.44}{0.67} & \avgstd{88.22}{1.46} & \avgstd{97.70}{0.13} & \avgstd{80.00}{1.18} & \avgstd{20.77}{0.40} \\
    & SSE & \avgstd{93.81}{0.42} & \avgstd{64.44}{2.47} & \avgstd{94.19}{1.32} & \avgstd{99.41}{0.13} & \avgstd{81.59}{0.29} & \avgstd{92.37}{0.78} & \avgstd{86.44}{0.44} & \avgstd{97.63}{0.13} & \avgstd{79.85}{2.86} & \avgstd{18.44}{1.52} \\
    & TTCov (ours) & \avgstd{94.70}{0.23} & \avgstd{68.74}{1.28} & \avgstd{95.07}{0.45} & \avgstd{99.56}{0.00} & \avgstd{80.88}{0.34} & \avgstd{93.63}{0.13} & \avgstd{89.11}{0.89} & \avgstd{97.78}{0.00} & \avgstd{78.81}{1.85} & \avgstd{20.62}{0.52} \\
    \midrule
    \multirow{4}{*}{0.75x} & Random & \avgstd{93.78}{0.51} & \avgstd{63.33}{3.01} & \avgstd{92.93}{0.64} & \avgstd{99.41}{0.13} & \avgstd{81.12}{0.86} & \avgstd{92.30}{0.78} & \avgstd{88.74}{0.34} & \avgstd{97.78}{0.00} & \avgstd{78.52}{0.93} & \avgstd{18.76}{1.36} \\
    & Coreset & \avgstd{94.67}{0.68} & \avgstd{74.67}{2.22} & \avgstd{97.22}{0.38} & \avgstd{99.33}{0.00} & \avgstd{82.87}{0.75} & \avgstd{93.63}{0.26} & \avgstd{91.04}{1.80} & \avgstd{97.63}{0.13} & \avgstd{77.93}{3.37} & \avgstd{21.48}{1.80} \\
    & SSE & \avgstd{94.81}{0.23} & \avgstd{72.37}{1.68} & \avgstd{96.89}{0.73} & \avgstd{99.33}{0.00} & \avgstd{81.43}{0.16} & \avgstd{92.96}{0.13} & \avgstd{88.81}{1.22} & \avgstd{97.63}{0.13} & \avgstd{78.81}{1.80} & \avgstd{22.29}{1.10} \\
    & TTCov (ours) & \avgstd{94.96}{0.76} & \avgstd{73.85}{3.90} & \avgstd{97.70}{0.65} & \avgstd{99.63}{0.13} & \avgstd{81.39}{0.64} & \avgstd{93.70}{0.90} & \avgstd{91.56}{0.80} & \avgstd{97.78}{0.00} & \avgstd{78.37}{1.36} & \avgstd{23.00}{1.53} \\
    \midrule
    \multirow{4}{*}{1x} & Random & \avgstd{94.33}{1.25} & \avgstd{68.15}{0.64} & \avgstd{94.96}{0.55} & \avgstd{99.48}{0.13} & \avgstd{81.69}{0.44} & \avgstd{92.52}{1.00} & \avgstd{89.11}{2.19} & \avgstd{97.78}{0.00} & \avgstd{79.41}{1.68} & \avgstd{20.15}{0.48} \\
    & Coreset & \avgstd{95.04}{0.74} & \avgstd{75.41}{2.58} & \avgstd{98.11}{0.69} & \avgstd{99.33}{0.00} & \avgstd{82.56}{0.12} & \avgstd{93.63}{0.13} & \avgstd{91.78}{1.33} & \avgstd{97.70}{0.13} & \avgstd{74.81}{7.12} & \avgstd{23.63}{1.19} \\
    & SSE & \avgstd{94.85}{0.23} & \avgstd{74.96}{2.38} & \avgstd{97.15}{0.74} & \avgstd{99.48}{0.13} & \avgstd{82.00}{0.83} & \avgstd{93.04}{0.51} & \avgstd{91.26}{1.73} & \avgstd{97.78}{0.00} & \avgstd{78.81}{3.22} & \avgstd{23.45}{0.53} \\
    & TTCov (ours) & \avgstd{95.04}{0.97} & \avgstd{78.07}{4.14} & \avgstd{98.30}{0.51} & \avgstd{99.48}{0.13} & \avgstd{82.05}{0.13} & \avgstd{94.22}{0.80} & \avgstd{92.44}{0.67} & \avgstd{97.78}{0.00} & \avgstd{78.37}{2.96} & \avgstd{24.42}{0.66} \\
    \midrule
    \multirow{4}{*}{1.25x} & Random & \avgstd{94.63}{1.05} & \avgstd{69.85}{2.36} & \avgstd{95.52}{0.90} & \avgstd{99.63}{0.13} & \avgstd{80.93}{0.01} & \avgstd{92.96}{1.26} & \avgstd{89.48}{0.26} & \avgstd{97.70}{0.13} & \avgstd{78.07}{1.03} & \avgstd{21.95}{0.42} \\
    & Coreset & \avgstd{95.70}{0.83} & \avgstd{76.81}{0.84} & \avgstd{98.44}{0.69} & \avgstd{99.33}{0.00} & \avgstd{82.94}{0.53} & \avgstd{94.22}{0.00} & \avgstd{93.56}{0.44} & \avgstd{97.63}{0.13} & \avgstd{79.70}{0.51} & \avgstd{24.55}{1.39} \\
    & SSE & \avgstd{95.78}{0.87} & \avgstd{76.00}{2.56} & \avgstd{97.78}{0.77} & \avgstd{99.48}{0.13} & \avgstd{81.78}{1.15} & \avgstd{94.22}{1.46} & \avgstd{91.85}{0.34} & \avgstd{97.70}{0.13} & \avgstd{79.56}{0.89} & \avgstd{24.76}{1.84} \\
    & TTCov (ours) & \avgstd{94.89}{0.51} & \avgstd{76.59}{2.63} & \avgstd{96.89}{1.28} & \avgstd{99.41}{0.13} & \avgstd{81.43}{0.35} & \avgstd{94.30}{0.71} & \avgstd{91.56}{2.40} & \avgstd{97.78}{0.00} & \avgstd{80.15}{4.38} & \avgstd{25.48}{0.58} \\
    \midrule
    \multirow{4}{*}{1.5x} & Random & \avgstd{94.74}{0.71} & \avgstd{75.04}{1.51} & \avgstd{96.52}{1.66} & \avgstd{99.48}{0.13} & \avgstd{81.47}{1.56} & \avgstd{93.63}{0.84} & \avgstd{90.22}{2.62} & \avgstd{97.78}{0.00} & \avgstd{73.78}{6.22} & \avgstd{22.95}{1.20} \\
    & Coreset & \avgstd{95.59}{0.36} & \avgstd{78.59}{1.45} & \avgstd{98.30}{0.13} & \avgstd{99.48}{0.26} & \avgstd{82.27}{0.27} & \avgstd{94.44}{0.38} & \avgstd{92.52}{3.56} & \avgstd{97.70}{0.13} & \avgstd{80.00}{0.44} & \avgstd{25.89}{0.50} \\
    & SSE & \avgstd{95.96}{0.45} & \avgstd{79.19}{1.89} & \avgstd{98.22}{0.69} & \avgstd{99.48}{0.13} & \avgstd{81.51}{0.62} & \avgstd{94.96}{0.46} & \avgstd{91.85}{1.30} & \avgstd{97.63}{0.13} & \avgstd{81.48}{0.26} & \avgstd{26.03}{1.22} \\
    & TTCov (ours) & \avgstd{94.67}{0.67} & \avgstd{78.59}{2.31} & \avgstd{98.70}{0.57} & \avgstd{99.33}{0.00} & \avgstd{82.85}{0.09} & \avgstd{93.48}{0.90} & \avgstd{93.85}{1.14} & \avgstd{97.78}{0.00} & \avgstd{80.59}{1.85} & \avgstd{26.40}{1.26} \\
    \midrule
    -- & Navtrain (oracle) & \avgstd{95.56}{0.79} & \avgstd{77.78}{2.83} & \avgstd{97.61}{0.39} & \avgstd{99.44}{0.16} & \avgstd{84.29}{0.33} & \avgstd{94.56}{0.16} & \avgstd{92.67}{2.83} & \avgstd{97.78}{0.00} & \avgstd{77.78}{1.89} & \avgstd{24.49}{1.41} \\
    \bottomrule
  \end{tabular}%
  }

  \vspace{0.2mm}
  \textbf{Stage 2}

  \vspace{0.2mm}

  \resizebox{\textwidth}{!}{%
  \begin{tabular}{@{\hspace{0.4em}}cccccccccccc@{\hspace{0.4em}}}
    \toprule
    Budget & Method & NC & DAC & DDC & TLC & EP & TTC & LK & HC & EC & EPDMS \\
    \midrule
    \multirow{4}{*}{0.5x} & Random & \avgstd{79.86}{0.87} & \avgstd{58.74}{0.22} & \avgstd{75.97}{1.81} & \avgstd{97.99}{0.10} & \avgstd{77.64}{0.38} & \avgstd{77.43}{1.46} & \avgstd{44.58}{2.02} & \avgstd{97.59}{0.02} & \avgstd{81.26}{1.62} & \avgstd{18.95}{0.39} \\
    & Coreset & \avgstd{79.63}{1.02} & \avgstd{62.92}{0.65} & \avgstd{77.70}{1.70} & \avgstd{98.02}{0.18} & \avgstd{80.28}{1.83} & \avgstd{77.01}{0.43} & \avgstd{42.89}{0.44} & \avgstd{97.71}{0.41} & \avgstd{79.91}{1.05} & \avgstd{20.77}{0.40} \\
    & SSE & \avgstd{80.04}{0.84} & \avgstd{60.03}{1.06} & \avgstd{75.94}{2.59} & \avgstd{97.96}{0.22} & \avgstd{78.52}{0.22} & \avgstd{78.01}{1.00} & \avgstd{44.28}{1.51} & \avgstd{97.01}{0.66} & \avgstd{81.59}{2.53} & \avgstd{18.44}{1.52} \\
    & TTCov (ours) & \avgstd{81.21}{0.63} & \avgstd{61.64}{1.05} & \avgstd{78.07}{0.93} & \avgstd{97.79}{0.48} & \avgstd{78.56}{1.19} & \avgstd{78.84}{0.62} & \avgstd{44.29}{0.16} & \avgstd{97.55}{0.37} & \avgstd{80.97}{2.16} & \avgstd{20.62}{0.52} \\
    \midrule
    \multirow{4}{*}{0.75x} & Random & \avgstd{80.54}{2.17} & \avgstd{61.38}{1.56} & \avgstd{76.55}{1.22} & \avgstd{98.16}{0.24} & \avgstd{77.70}{2.28} & \avgstd{77.97}{1.76} & \avgstd{43.41}{1.26} & \avgstd{97.35}{0.29} & \avgstd{80.72}{2.83} & \avgstd{18.76}{1.36} \\
    & Coreset & \avgstd{79.15}{0.66} & \avgstd{65.37}{1.81} & \avgstd{79.88}{1.80} & \avgstd{97.72}{0.49} & \avgstd{82.11}{0.38} & \avgstd{76.08}{1.27} & \avgstd{45.52}{0.35} & \avgstd{97.03}{0.32} & \avgstd{77.89}{1.40} & \avgstd{21.48}{1.80} \\
    & SSE & \avgstd{81.39}{1.15} & \avgstd{62.97}{0.73} & \avgstd{78.23}{0.85} & \avgstd{97.85}{0.22} & \avgstd{80.23}{0.89} & \avgstd{78.59}{1.39} & \avgstd{44.29}{0.73} & \avgstd{97.12}{0.30} & \avgstd{80.18}{0.73} & \avgstd{22.29}{1.10} \\
    & TTCov (ours) & \avgstd{80.05}{0.44} & \avgstd{66.78}{2.05} & \avgstd{82.08}{1.93} & \avgstd{98.14}{0.11} & \avgstd{80.39}{0.89} & \avgstd{77.28}{0.45} & \avgstd{45.42}{2.75} & \avgstd{97.56}{0.67} & \avgstd{78.75}{0.84} & \avgstd{23.00}{1.53} \\
    \midrule
    \multirow{4}{*}{1x} & Random & \avgstd{80.86}{1.28} & \avgstd{62.52}{1.91} & \avgstd{78.41}{2.39} & \avgstd{97.89}{0.58} & \avgstd{80.19}{0.99} & \avgstd{78.53}{1.76} & \avgstd{44.91}{3.43} & \avgstd{97.04}{0.40} & \avgstd{80.29}{2.24} & \avgstd{20.15}{0.48} \\
    & Coreset & \avgstd{80.29}{0.91} & \avgstd{67.33}{2.68} & \avgstd{81.43}{0.93} & \avgstd{97.94}{0.19} & \avgstd{81.63}{0.18} & \avgstd{78.13}{1.54} & \avgstd{46.04}{1.44} & \avgstd{97.03}{0.33} & \avgstd{71.79}{5.99} & \avgstd{23.63}{1.19} \\
    & SSE & \avgstd{81.24}{1.06} & \avgstd{67.08}{0.27} & \avgstd{80.65}{1.32} & \avgstd{98.19}{0.20} & \avgstd{80.51}{0.94} & \avgstd{79.24}{0.86} & \avgstd{45.10}{1.17} & \avgstd{97.18}{0.37} & \avgstd{77.69}{0.60} & \avgstd{23.45}{0.53} \\
    & TTCov (ours) & \avgstd{80.80}{0.20} & \avgstd{69.03}{0.95} & \avgstd{81.85}{0.58} & \avgstd{97.73}{0.37} & \avgstd{81.36}{0.50} & \avgstd{77.76}{1.58} & \avgstd{43.94}{0.05} & \avgstd{97.03}{0.35} & \avgstd{78.79}{1.11} & \avgstd{24.42}{0.66} \\
    \midrule
    \multirow{4}{*}{1.25x} & Random & \avgstd{80.94}{0.81} & \avgstd{66.51}{1.13} & \avgstd{80.86}{1.22} & \avgstd{98.08}{0.23} & \avgstd{78.61}{1.16} & \avgstd{78.83}{1.66} & \avgstd{44.70}{1.49} & \avgstd{96.88}{0.07} & \avgstd{79.00}{3.13} & \avgstd{21.95}{0.42} \\
    & Coreset & \avgstd{80.05}{0.88} & \avgstd{70.54}{0.63} & \avgstd{83.41}{0.63} & \avgstd{98.05}{0.53} & \avgstd{82.53}{0.87} & \avgstd{77.10}{1.81} & \avgstd{48.22}{1.46} & \avgstd{97.21}{0.13} & \avgstd{74.45}{1.74} & \avgstd{24.55}{1.39} \\
    & SSE & \avgstd{80.44}{0.62} & \avgstd{70.51}{2.50} & \avgstd{82.80}{1.53} & \avgstd{98.30}{0.24} & \avgstd{81.26}{0.72} & \avgstd{78.17}{1.03} & \avgstd{47.05}{0.59} & \avgstd{96.58}{0.37} & \avgstd{77.51}{2.21} & \avgstd{24.76}{1.84} \\
    & TTCov (ours) & \avgstd{80.90}{1.74} & \avgstd{69.60}{1.53} & \avgstd{83.24}{0.90} & \avgstd{98.29}{0.49} & \avgstd{80.65}{0.81} & \avgstd{78.37}{1.82} & \avgstd{47.58}{0.70} & \avgstd{97.43}{0.24} & \avgstd{77.11}{0.91} & \avgstd{25.48}{0.58} \\
    \midrule
    \multirow{4}{*}{1.5x} & Random & \avgstd{80.13}{0.91} & \avgstd{67.63}{2.03} & \avgstd{80.83}{0.59} & \avgstd{97.90}{0.36} & \avgstd{80.22}{2.38} & \avgstd{78.18}{1.08} & \avgstd{45.25}{1.00} & \avgstd{97.08}{0.10} & \avgstd{73.95}{1.89} & \avgstd{22.95}{1.20} \\
    & Coreset & \avgstd{81.53}{0.40} & \avgstd{70.53}{0.26} & \avgstd{82.53}{1.25} & \avgstd{98.09}{0.14} & \avgstd{82.35}{0.57} & \avgstd{78.76}{0.72} & \avgstd{47.77}{0.61} & \avgstd{97.04}{0.20} & \avgstd{74.86}{0.75} & \avgstd{25.89}{0.50} \\
    & SSE & \avgstd{80.35}{1.84} & \avgstd{69.23}{0.37} & \avgstd{82.63}{0.39} & \avgstd{98.46}{0.34} & \avgstd{82.11}{1.23} & \avgstd{78.77}{1.16} & \avgstd{45.14}{1.02} & \avgstd{96.76}{0.35} & \avgstd{76.89}{1.82} & \avgstd{26.03}{1.22} \\
    & TTCov (ours) & \avgstd{80.19}{0.29} & \avgstd{69.33}{1.03} & \avgstd{83.08}{0.41} & \avgstd{98.13}{0.57} & \avgstd{82.32}{0.13} & \avgstd{77.64}{0.58} & \avgstd{46.06}{0.75} & \avgstd{96.77}{0.14} & \avgstd{76.29}{1.07} & \avgstd{26.40}{1.26} \\
    \midrule
    -- & Navtrain (oracle) & \avgstd{80.20}{1.25} & \avgstd{66.93}{1.28} & \avgstd{80.76}{0.04} & \avgstd{98.58}{0.10} & \avgstd{85.45}{0.21} & \avgstd{77.95}{1.60} & \avgstd{44.46}{0.08} & \avgstd{96.39}{0.85} & \avgstd{72.65}{0.55} & \avgstd{24.49}{1.41} \\
    \bottomrule
  \end{tabular}%
  }
  \vspace{0.2em}

  \parbox{\textwidth}{\scriptsize\raggedright \textbf{Abbreviations:}
  NC = no at-fault collisions; DAC = drivable area compliance;
  DDC = driving direction compliance; TLC = traffic light compliance;
  EP = ego progress; TTC = time-to-collision within bound; LK = lane keeping;
  HC = history comfort; EC = extended comfort.}
  \vspace{-2mm}
  \endgroup
\end{table*}

\section{Optimizer Ablations}
\label{app:optim}

\subsection{Greedy Residual Matching}
A natural alternative to KL distribution matching is to match AP counts
directly. Let $\mathbf{n}(x)$ denote the AP-count vector of sample $x$, with
entries $n(x, L)$. Define the desired AP totals at budget $B$ as
\begin{equation} \label{eq:rm_target}
    \mathbf{m} \;=\; B\,p^\star.
\end{equation}
If $\mathcal{S}^{(t)}$ is the selected set after $t$ iterations, define the
residual
\begin{equation} \label{eq:rm_residual}
    \mathbf{r}^{(t)} \;=\; \mathbf{m} \;-\; \sum_{x \in \mathcal{S}^{(t)}} \mathbf{n}(x).
\end{equation}
Using weighted squared error,
\begin{equation} \label{eq:rm_loss}
    L(\mathcal{S}) \;=\; \left\| \mathbf{m} \;-\; \sum_{x \in \mathcal{S}} \mathbf{n}(x) \right\|_W^2,
    \qquad W = \mathrm{diag}(w_1, \ldots).
\end{equation}
If candidate sample $x$ is added, the new residual is
$\mathbf{r}^{(t)} - \mathbf{n}(x)$, and the improvement is
\begin{equation} \label{eq:rm_improvement}
    \|\mathbf{r}^{(t)}\|_W^2 \;-\; \|\mathbf{r}^{(t)} - \mathbf{n}(x)\|_W^2
    \;=\; 2\,\mathbf{n}(x)^\top W \mathbf{r}^{(t)} \;-\; \mathbf{n}(x)^\top W \mathbf{n}(x).
\end{equation}
Therefore the greedy step is
\begin{equation} \label{eq:rm_step}
    x^{(t+1)} \;=\; \arg\max_{x \,\in\, \mathcal{C} \setminus \mathcal{S}^{(t)}}
    \left\{\, 2\,\mathbf{n}(x)^\top W \mathbf{r}^{(t)} \;-\; \mathbf{n}(x)^\top W \mathbf{n}(x) \,\right\}.
\end{equation}
The first term rewards samples that fill APs we still need. The second
penalizes any AP added, so as the residual
decreases, we naturally stops adding samples that would overshoot. We use the
smoothed weighting $w_L = 1/(p^\star(L) + \tau)$ with $\tau = 1/B$ as the
default, which uplifts rare APs without diverging as $p^\star(L) \to 0$. Setting $W = I$ recovers the unweighted variant.

\subsection{Repeat Factor Sampling}

The proportional K-Atlas target $p^\star$ in \eqref{eq:p_star_L} contains many common APs, so a budget constrained optimization can under select rare APs. We adapt repeat-factor sampling (RFS)~\citep{gupta2019lvis}, originally introduced for long-tailed instance segmentation, to uplift rare APs in the target before optimization. Let $c(L)$ denote the raw Atlas count for AP $L$ and $N_{\mathrm{test}}$ the number of test samples, so the total AP frequency is
$f(L) = c(L) / N_{\mathrm{test}}$. We compute a repeat factor per AP
\begin{equation} \label{eq:rfs_factor}
    r(L) \;=\; \max\!\left(1,\; \sqrt{t\,/\,f(L)}\right),
\end{equation}
where $t$ is the threshold hyperparameter (default $t = 10^{-2}$). APs with
$f(L) \geq t$ are unchanged ($r(L) = 1$), and rarer APs are amplified inversely
to the square root of their frequency. The reweighted target distribution is
\begin{equation} \label{eq:rfs_target}
    p_{\mathrm{RFS}}(L) \;=\; \frac{c(L)\, r(L)}{\sum_{L'} c(L')\, r(L')},
\end{equation}
which substitutes for $p^\star$ in \eqref{eq:ttcov_opt_problem}. Intuitively,
$t$ controls how aggressively rare APs get amplified. At low $t$ (\eg $10^{-3}$)
only the rarest APs are boosted and the target stays close to proportional,
while at high $t$ (\eg $10^{-1}$) most APs count as rare and end up taking a larger share of the budget than they did under proportional weighting.

\subsection{Penalty}
To approach \eqref{eq:ttcov_opt_problem} greedily, we add one sample at a time.
Let $\mathcal{S}^{(t)}$ be the curated set after $t$ picks, $n(x, L)$ the count
of AP $L$ in sample $x$, $n(x) = \sum_L n(x, L)$, and
$q^{(t)}(L) = \sum_{x \in \mathcal{S}^{(t)}} n(x, L)$ with $S^{(t)} = \sum_L q^{(t)}(L)$
the running coverage counts and total mass. Expanding the marginal KL gain
$D_{\mathrm{KL}}(p^\star \,\|\, \hat{p}_{\mathcal{S}^{(t)}}) - D_{\mathrm{KL}}(p^\star \,\|\, \hat{p}_{\mathcal{S}^{(t+1)}})$
gives the per-candidate score
\begin{equation} \label{eq:greedy_kl_original}
    \mathrm{score}(x)
    \;=\;
    \sum_{L} p^\star(L)\,\log\!\left(\frac{q^{(t)}(L) + n(x, L)}{q^{(t)}(L)}\right)
    \;-\;
    \log\!\left(\frac{S^{(t)} + n(x)}{S^{(t)}}\right),
\end{equation}
and we pick $x^{(t+1)} = \arg\max_{x \in \mathcal{C} \setminus \mathcal{S}^{(t)}} \mathrm{score}(x)$.

Equation \eqref{eq:greedy_kl_original} treats every AP symmetrically, so a
sample whose APs would only added to saturated K-Atlas APs
$B\,p^\star(L)$ is scored only through its KL gain, not its overshoot. We
introduce a penalty $\rho \geq 0$ that penalizes candidates by how much they add to already saturated APs:
\begin{equation} \label{eq:greedy_kl_penalty}
    \mathrm{score}^{(\rho)}(x)
    \;=\;
    \mathrm{score}(x)
    \;-\;
    (\rho - 1)\,\frac{\sum_{L} n(x, L)\,\bigl[q^{(t)}(L) - B\,p^\star(L)\bigr]_+}{S^{(t)}},
\end{equation}
with $[z]_+ = \max(0, z)$. Setting $\rho = 1$ zeroes the second term, recovering our original greedy score
\eqref{eq:greedy_kl_original}. $\rho > 1$ penalizes over-coverage and
$\rho < 1$ rewards it.

\section{Datasets and Virtual Clip Creation}
\label{app:data}
In this section, we provide details of the datasets and the virtual clip strategy we use for the experiments.  

The OpenScene dataset consists of continuous driving logs sampled at 2Hz. Frameworks such as Navsim typically process these logs using a sliding window to create individual "scenes". In Navsim, by default configuration, a scene is a 14-frame window consisting of a 3-frame history, 1 frame for the current state and a 10-frame future horizon. To better capture the actions during driving and align with both academic and industry standards~\cite{dimlioglu2026scaling}, we define a 20-frame virtual clip (10 seconds) as our fundamental unit for captioning and curation. Each 20-frame virtual clip therefore encapsulates 7 overlapping scenes. To ensure that every possible scene in the OpenScene trainval split is covered by one virtual clip, we employ a 7-frame stride to get the virtual clips. Such a virtual clip strategy makes sure all the scenes in OpenScene trainval are covered without the massive overhead of frame-by-frame sliding window. During the data curation process, we use the virtual clip as the fundamental unit, such that when a virtual clip is selected, all its 7 encapsulated scenes are added into the train dataset. 

\section{Additional Visualizations}
\begin{figure}
    \centering
    \includegraphics[width=1.0\linewidth]{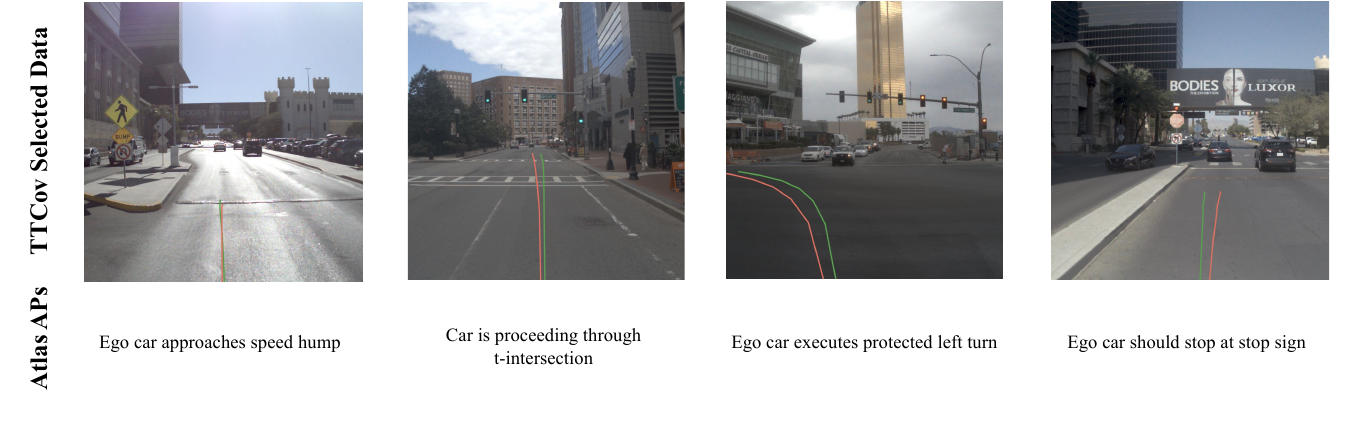}
    \caption{Additional visualization examples of TTCov selected data.}
    \label{fig:app_viz}
\end{figure}

In~\cref{fig:app_viz}, we provide additional visualizations of TTCov's selected data, demonstrating diversity in visual, semantic, and behavioral space.

\section{Code Release} All code will be released alongside the final version of this work.

\clearpage



\end{document}